\pdfoutput=1
\documentclass[11pt]{article}

\usepackage[final]{acl}

\usepackage{times}
\usepackage{latexsym}

\usepackage[T1]{fontenc}
\usepackage[utf8]{inputenc}

\usepackage{microtype}

\usepackage{inconsolata}

\usepackage{graphicx}
\usepackage{multirow}
\usepackage{subcaption}
\usepackage{physics}
\usepackage{amsmath}
\usepackage{amssymb}

\title{Householder Pseudo-Rotation: A Novel Approach to Activation Editing in LLMs with Direction-Magnitude Perspective}

\author{Van-Cuong Pham \\
  VinAI Research \\
  Hanoi, Vietnam \\
  \texttt{v.cuongpv27@vinai.io} \\\And
  Thien Huu Nguyen \\
  University of Oregon \\ Department of Computer Science \\ Eugene, OR, USA \\
  \texttt{thien@cs.uoregon.edu} \\}

\begin{document}
\maketitle
\begin{abstract}
Activation Editing, which involves directly editing the internal representations of large language models (LLMs) to alter their behaviors and achieve desired properties, has emerged as a promising area of research. Existing works primarily treat LLMs' activations as points in space and modify them by adding steering vectors. However, this approach is limited in its ability to achieve greater performance improvement while maintaining the necessary consistency of activation magnitudes. To overcome these issues, we propose a novel editing method that views activations in terms of their directions and magnitudes. Our method, named \emph{Householder Pseudo-Rotation} (HPR), mimics the rotation transformation, thus preserving activation norms and resulting in an improved performance on various safety benchmarks.% We release our source code at <Input github link here>.

%We show that doing so would break the magnitude consistency of the activation vectors in LLMs. To overcome this shortcoming, we propose a novel editing method that views activations in terms of their directions and magnitudes. Our method, which we name \emph{Householder Pseudo-Rotation} (HPR), mimics the rotation transformation, thus preserving activation norms and resulting in an improved performance on various safety benchmarks.
\end{abstract}

\section{Introduction}
\label{sec:introduction}

Building upon the paradigm of pre-training language models on large corpora of raw text using next-sentence-prediction objective (\citealp[]{Radford2018ImprovingLU}; \citealp[]{radford2019language}), Large Language Models (LLMs) research has taken a big leap and become an essential asset of AI in recent years. The latest LLMs can exhibit phenomenal fluency and reasoning capability, excel in numerous NLP benchmarks, while also aligning to human intent (\citealp[]{wei2022emergent}; \citealp[]{NEURIPS2022_b1efde53}; \citealp[]{touvron2023llama}; \citealp[]{jiang2023mistral}; \citealp[]{openai2024gpt4}). In the midst of the rapid development of LLMs, efforts to study and control their societal impacts, including issues such as hallucination, bias, and toxicity to name a few, are of the utmost importance. Yet, with their ever-growing size, reaching hundreds of billions of parameters (\citealp[]{NEURIPS2020_1457c0d6}; \citealp[]{chowdhery2022palm}), the popular approach for controlling and aligning LLMs via fine-tuning proves to be very challenging and resource-intensive, necessitating the research into alternative solutions to adapt the behaviors of LLMs.

%their pre-trained knowledge. 

%LLMs prove to be very challenging and resource-intensive to fine-tune, necessitating the research into alternative solutions to elicit their pre-trained knowledge. 

\begin{figure*}[hbt!]
    \centering
    \begin{subfigure}{0.31\textwidth}
        \includegraphics[width=\linewidth]{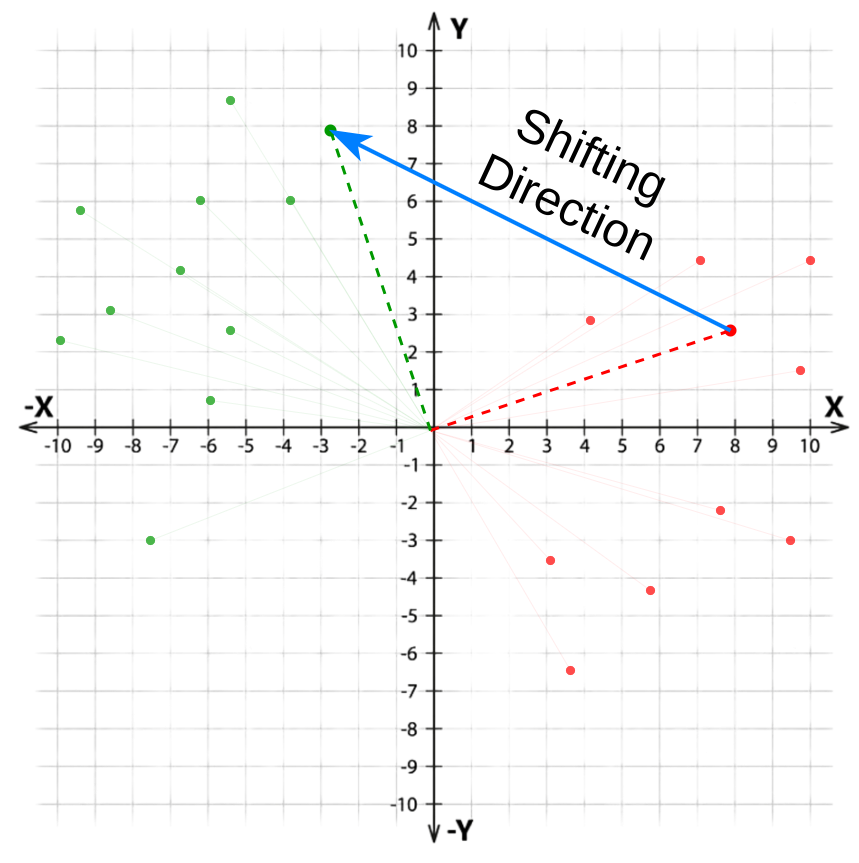}
        \caption{Points-in-space view}
        \label{fig:views-a}
    \end{subfigure}
    % \hspace*{\fill}
    \begin{subfigure}{0.31\textwidth}
        \includegraphics[width=\linewidth]{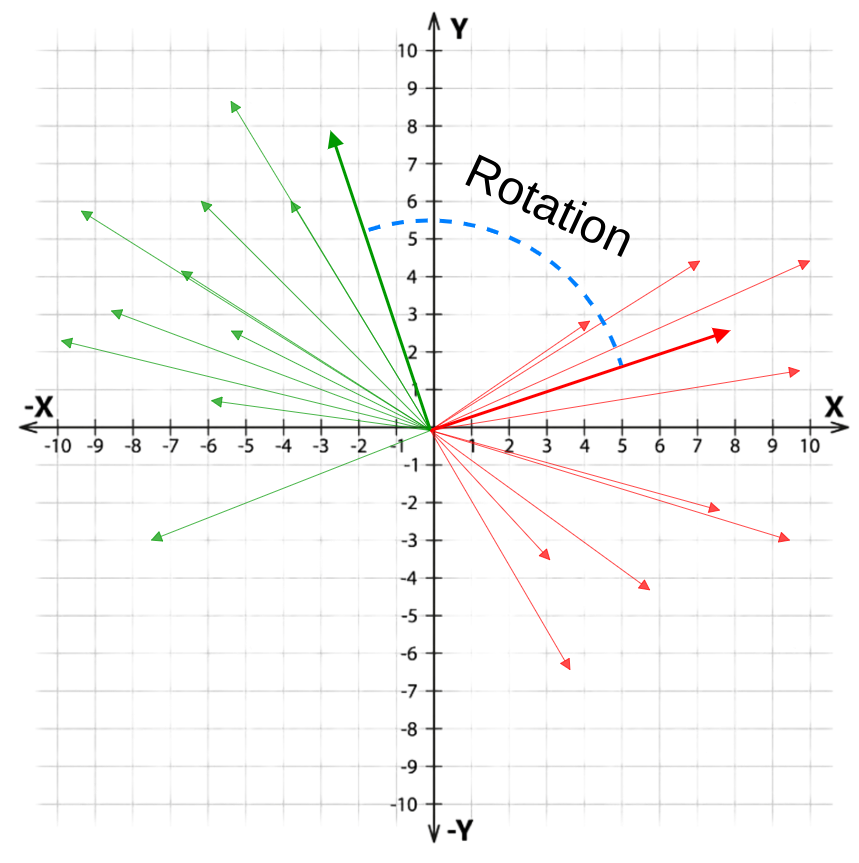}
        \caption{Direction-magnitude view}
        \label{fig:views-b}
    \end{subfigure}
    % \hspace*{\fill}
    \begin{subfigure}{0.31\textwidth}
        \includegraphics[width=\linewidth]{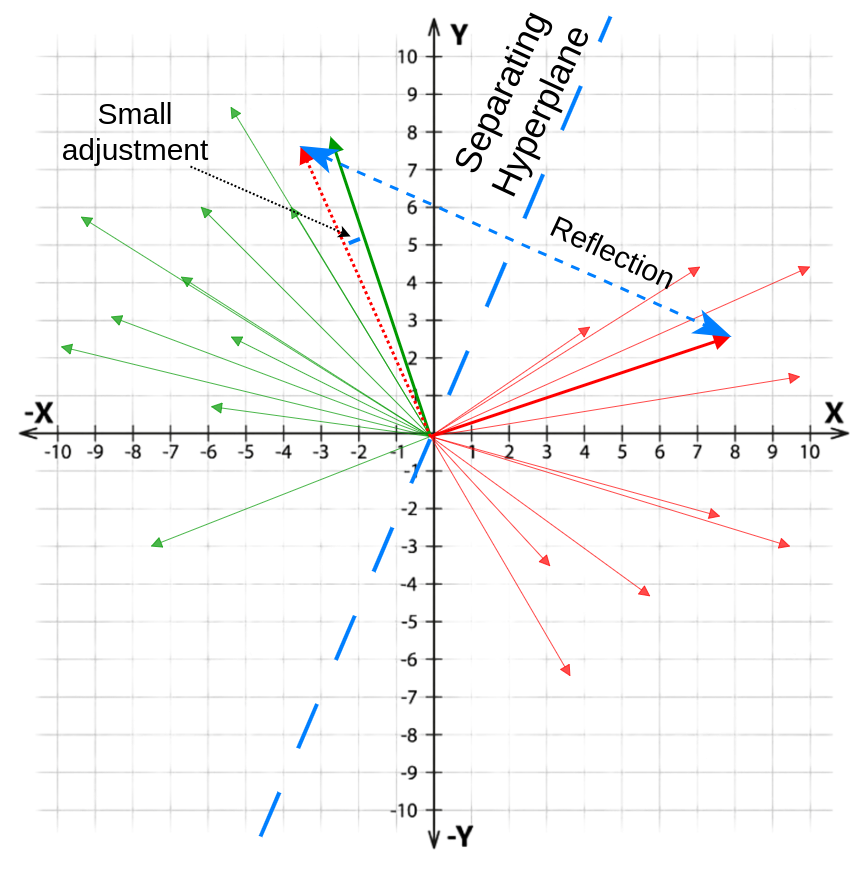}
        \caption{Our approach}
        \label{fig:views-c}
    \end{subfigure}
    % \hspace*{\fill}
  
    % \centering
    % \includegraphics[width=0.48\linewidth]{images/reflection.drawio.png}
  \caption {Comparison of points-in-space view (a) and direction-magnitude view (b). Positive activations are colored \textcolor{green}{green} and negative activations are colored \textcolor{red}{red}. The editing methods are depicted in in \textcolor{blue}{blue}. Our proposed method (c) approximates the rotation transformation by first reflecting negative activations through a learned separating hyperplane and then adjusting the reflections to reach the right angle.}
  \label{fig:views}
\end{figure*}

Among various approaches to efficiently adapt LLMs (\citealp[]{lester-etal-2021-power}; \citealp[]{li-liang-2021-prefix}; \citealp[]{hu2022lora}; \citealp[]{dong2023survey}; \citealp[]{wan2024efficient}), Activation Editing, also referred to as ``Intervention'' or ``Representation Engineering'' in the literature, has shown promising results. Based on the observation that LLMs form an internal ``belief'' about facts in their activation space even before the responses are generated (\citealp[]{dai-etal-2022-knowledge}; \citealp[]{NEURIPS2023_81b83900}; \citealp[]{burns2023discovering}; \citealp[]{joshi2024personas}), this approach aims to draw factual knowledge out of the model by directly editing activation vectors at inference time. Most existing works in this area utilize a \emph{steering vector} (\citealp[]{NEURIPS2023_81b83900}; \citealp[]{turner2023activation}, \citealp[]{rimsky2024steering}; \citealp[]{vonrütte2024language}), which can be scaled by a scaling factor and added to the original activation. In doing so, activations are viewed as \emph{points in space} (Figure \ref{fig:views-a}). Correspondingly, the process of adding a fixed steering vector to activations can be interpreted as moving these points along a vector offset \cite{mikolov-etal-2013-linguistic}, and the scaling factor tells how far they should be moved. 

\iffalse 
In an experiment with the activation space we find that an LLM's layer has two important properties, which we will discuss further in Section \ref{sec:analysis_act}: 1) \textbf{Magnitude consistency}: Activations of the same layer tend to have roughly the same vector norm; and 2) \textbf{Magnitude escalation}: The activation norms increase exceedingly as the layer gets deeper.
\fi 

In an experiment with the activation space, we discover an important property that is maintained by powerful LLMs: activations within the same layer tend to have roughly the same vector norm. We refer to this as the \textbf{Magnitude Consistency} property, i.e., Section \ref{sec:analysis_act}. This observation highlights a key limitation of the points-in-space view, where the steering vector approach cannot simultaneously maintain activation magnitude consistency and effectively edit activation to achieve greater performance improvement for desired behaviors for LLMs. If the scaling factor is too large, the additive edit might drastically alter the activation norms in each layer, violating the norm consistency property of LLMs. In extreme cases, this change can lead to the generation of complete gibberish, undermining the desired behaviors of the LLM's responses. Conversely, if the scaling factor is set too low to preserve the activation norms, the steering vector may have limited abilities to shift an activation toward new behavior, thus also hindering editing performance for desired behaviors. Moreover, the steering vector approach does not align with the commonly used cosine similarity metric, which emphasizes directional alignment between vectors rather than their absolute positions.

%This observation reveals problematic aspects of the aforementioned points-in-space view.

%without a carefully designed scaling factor and steering vector, the additive edit might change the activation norms in each layer drastically, thus violating this property. In extreme cases, this change in activation norms can lead to the generation of complete gibberish, which hinders the desired behaviors of the responses from LLMs. Second, this view does not conform to the commonly used cosine similarity metric, which focuses on the directional alignment between vectors rather than their absolute positions.

We argue that activation vectors should instead be understood in terms of their directions and magnitudes. We call this the \emph{direction-magnitude} view (Figure \ref{fig:views-b}). In this regard, the semantic information of activations is reflected in their directions from the origin, while their magnitude represents the intensity of such information. This view also facilitates cosine similarity better since it measures the relationship between activations via the angle between their directions. Furthermore, while the points-in-space view struggles to achieve activation norm consistency, the direction-magnitude view can conveniently interpret the activation space in each layer as a $(d-1)$-dimensional hypersphere centered at the origin. As such, the activations can have a ``stable'' norm via the sphere's radius. \iffalse (magnitude consistency), and a sphere corresponding to a deeper layer should have a larger radius than shallow layers' (magnitude escalation). \fi

In this work, we introduce a novel editing method based on the direction-magnitude view. Instead of trying to move points, our method aims to alter a LLM's behavior by rotating activation vectors around the origin to their designated directions (Figure \ref{fig:views-b}). For example, rotating from untruthful region into truthful region. Usually, computing a matrix for vector rotation is non-trivial, especially in high-dimensional space. Therefore, we propose to relax the problem and resort to an approximated rotation transformation instead (Figure \ref{fig:views-c}). To this end, we first determine a hyperplane going through the origin that separates the two regions of interest. We then reflect undesirable activations about this hyperplane to make them land on the desirable region. Having an unique hyperplane for each individual activation vector is infeasible computationally as it would cost substantial GPU memory to store them at runtime. We thus learn a global hyperplane separating the activation vectors for each edited layer. Finally, for each reflection of an undersriable activation, we adjust it to the corresponding desired activation. In this way, our solution is more efficient as the adjustment for each activation only involves scalar angles, whose learning is less expensive than a rotation matrix for each edited vector. We name this method \emph{Householder Pseudo-Rotation} (HPR), based on the Householder transformation \cite{10.1145/320941.320947} at its core.

We evaluate our editing method HPR on eliciting truthfulness from LLMs. Experiment results on the TruthfulQA dataset \cite{lin-etal-2022-truthfulqa} demonstrate a significant boost in performance compared to Steering Vector editing. We further show that HPR can improve LLMs' performance for other behavior-related problems, including bias, ethics, and toxicity. Finally, we conduct extensive analysis to provide deeper insights for the advantages of HPR for activation editing.

%Following previous work, we focus on eliciting truthfulness from LLM. Experiment results on TruthfulQA dataset \cite{lin-etal-2022-truthfulqa} demonstrate a significant boost in performance compared to Steering Vector editing. We further show that HPR can work for other behavior-related problems, namely bias, ethics, and toxicity as well.

%Having stated the motivation \iffalse and scope of research\fi, we summarise our main contributions as follows:
%\begin{itemize}
%    \item We provide an analysis of activation vector norms in LLMs and highlight an important property of the activation space, which is \textbf{Magnitude Consistency}\iffalse and \textbf{magnitude escalation}\fi, and show how Steering Vector approach in Activation Editing may break this property, thereby hurting performance.
%    \item We propose a novel editing method based on the direction-magnitude view. This method, Householder Pseudo-Rotation, can preserve the consistency of LLMs' activation space, thus helping improve the model's behavior and outperform Steering Vector approach on various benchmarks.
% Fall short
%\end{itemize}

\section{Prerequisites}
\label{sec:prerequisites}

\subsection{Problem Statement}
\label{sec:notations}
Let $\mathcal{M} = \{\mathcal{M}^{(l)} | 0 \le l < L \}$ be a $L$-layers pre-trained LLM whose behavior needs to be altered. Assume that the outputs of $\mathcal{M}$ exhibit either of the two contrasting qualities: a positive behavior $\mathbf{p}$ or a negative behavior $\mathbf{n}$. For instance, $\mathbf{p}$ and $\mathbf{n}$ can be truthfulness and untruthfulness. We denote:

%\begin{itemize}
%    \item 
    
    $\bullet$ $x_i = \{x_{i,j} | 0 \le j < S^x\}$ : An input sequence of length $S^x$.
    
%    \item 
    
    $\bullet$ $y^{\mathbf{p}}_i = \{y^\mathbf{p}_{i, j} | 0 \le j < S^\mathbf{p}\}$ : The positive (i.e. desirable) output sequence with length $S^\mathbf{p}$.
    
%    \item 
    
    $\bullet$ $y^\mathbf{n}_i = \{y^\mathbf{n}_{i, j} | 0 \le j < S^\mathbf{n}\}$ : The negative (i.e. undesirable) output sequence with length $S^\mathbf{n}$.
    
%\end{itemize}
Here, $i$ is the sample index in the dataset, and $j$ is the token index in a sample. When the label of the output, i.e. positive or negative, is unknown, we refer to its length as $S^y$.

\iffalse
Our goal is to encourage $\mathcal{M}$ to generate $y^\mathbf{p}_i$ when given an arbitrary $x_i$. In other words, we want to maximize $\prod_{j=0}^{S^\mathbf{p}} p(y^\mathbf{p}_{i, j} | x_i, y_{i, <j})$, where $y_{i, <j} = \{y_{i, 0}, y_{i, 1}, \ldots, y_{i, j-1}\}$.
\fi

%and at token position $j$

In this work, unless specified otherwise,  a ``vector'' is understood as a column vector of size $d \times 1$. Let us further use $a^{\mathbf{p}, (l)}_{i, j} \in \mathcal{A}^{\mathbf{p}, (l)}$ to denote the $d$-dimensional positive activation vector at the $j^{th}$ token of the $l^{th}$ layer in $\mathcal{M}$, where $\mathcal{A}^{\mathbf{p}, (l)} \subset \mathbb{R}^d$ is the positive region in the activation space of $\mathcal{M}^{(l)}$. Similarly, the corresponding negative activation is denoted as $a^{\mathbf{n}, (l)}_{i, j} \in \mathcal{A}^{\mathbf{n}, (l)}$. These are obtained by forwarding the concatenation of the input and the corresponding output sequence, i.e. $x_i \Vert y_i^\mathbf{p}$ or $x_i \Vert y_i^\mathbf{n}$, through $\mathcal{M}$. Since the question part $x_i$ is the same for each data pair, we only use the activation vectors at the token positions of the responses. Without loss of generality, we omit the layer notation $(l)$ and the quality notation ($\mathbf{p}$ or $\mathbf{n}$) when referring to an arbitrary item. 

The general framework of Activation Editing utilizes an editing function $f(\cdot | \theta)$ with parameter $\theta$ for activation vectors $a_{i, j}$ such that $f(a_{i, j} | \theta) \in \mathcal{A}^\mathbf{p}$. The design of an Activation Editing method can thus be broken down to the the design of such a function and how to find the optimal $\theta$. For example, in Steering Vector methods \cite{NEURIPS2023_81b83900}, the editing function is a simple vector addition: $f(a_{i, j} | \theta) = a_{i, j} + \alpha \theta$ where $\alpha$ is a hyperparameter for scaling factor. 

\subsection{Householder Transformation}
\label{householder}

The idea of Householder transformation stemmed from an important lemma in \citet{10.1145/320941.320947} which stated: For any vector $a \ne 0$, and any unit vector $v$, there exists a unit vector $u$ such that:
\begin{equation}
    \centering
    (I - 2uu^T)a = \lVert a \rVert v
    \label{eq:householder_lemma}
\end{equation}

In this case, $\lVert a \rVert v$ is the reflection of $a$ about a hyperplane which passes through the origin and has $u$ as its normal vector. Since $v$ is a unit vector, $a$ and $\lVert a \rVert v$ have the same vector norm. Therefore, we can extend the problem to a more general case: For any pair of vectors $(a, b)$ of the same magnitude, it is possible to find a vector $c \ne 0$ such that:
\begin{equation}
    \centering
    b = (I - \frac{2cc^T}{c^Tc})a
    \label{eq:householder_re}
\end{equation}

The orthogonal matrix $H = (I - \frac{2cc^T}{c^Tc})$ is called the \emph{Householder Matrix}.

\section{Householder Pseudo-Rotation (HPR)}
\label{sec:hpr}

As discussed in the introduction, our goal is to find an editing function $f$ to alter the behavior of LLMs that can: 1) transform any vector in the activation spaces into one invoking positive behavior; 2) closely mimic the rotation transformation to preserve norm of the activations. The usual calculation of a rotation matrix between two $d$-dimensional vectors consists of several computationally expensive steps such as the Gram-Schmidt process, whose complexity is $\mathcal{O}(d^3)$. The Householder transformation (Equation \ref{eq:householder_re}) can be a cheaper alternative since it also retains the vector norm. However, in the context of Activation Editing, having a Householder matrix of size $d \times d$ for each activation vector would introduce too much extra data to be stored on GPU RAM, thus limiting applicability.

To alleviate these problems, we propose \emph{Householder Pseudo-Rotation} (HPR), a pseudo-rotation method that reflect negative activations in each layer about a global separating hyperplane and then adjust the resulting vectors to achieve the desired angle. The original problem is essentially broken down into two sub-problems: finding the separating hyperplane, and finding the rotating angle. We tackle them by incorporating into each edited layer a \textbf{linear probe} and an \textbf{angle prediction} module.

\subsection{Linear Probe}
\label{sec:hpr_linear_probe}
\begin{figure}[hbt!]
    \centering
    \includegraphics[width=0.9\columnwidth]{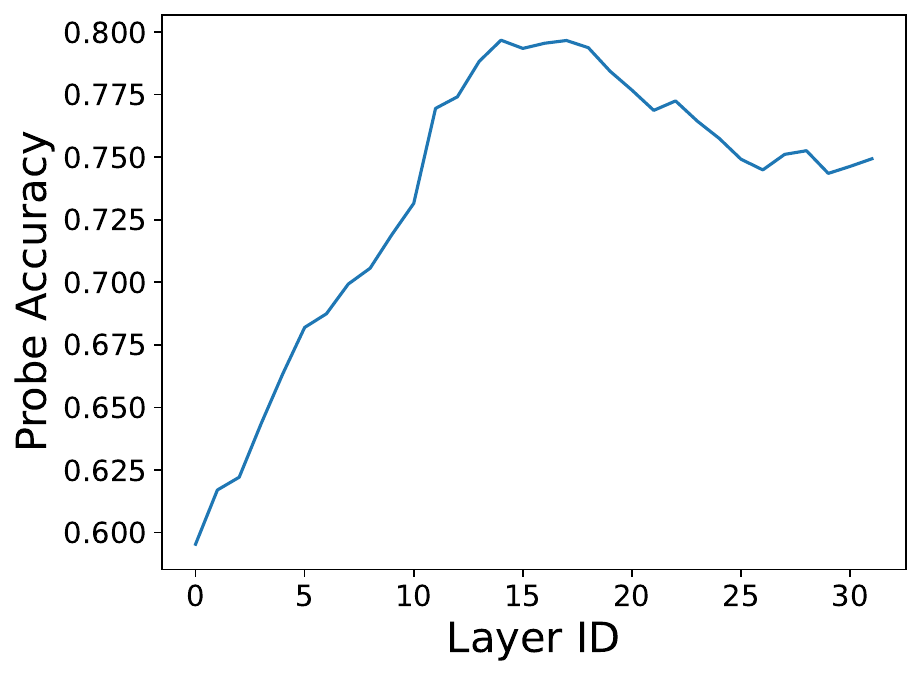}
    \caption{Probe accuracy of HPR-edited Llama2-7B-Chat on TruthfulQA. A linear probe is trained for each layer using positive-negative pairs of the training data and then evaluated on the validation data. \iffalse More details on TruthfulQA data splits can be found in Section \ref{sec:exp_setup}. The probes achieve non-trivial accuracy while reaching their peak at the middle layers.\fi}
    \label{fig:probe_acc_llama2}
\end{figure}

In the first step, we train a linear probe to discriminate the positive and negative activations of LLMs in each layer. The non-trivial accuracy of this probe, as can be seen in Figure \ref{fig:probe_acc_llama2}, suggests that it can effectively form a separating hyperplane between the positive and negative regions, and its weight vector serves as the normal vector of this hyperplane. We can then utilize the Householder matrix corresponding to this hyperplane as a means to reflect activations from one region to the other.

More concretely, the linear probe corresponding to a LLM layer can be defined as:
\begin{equation}
    \centering 
    f_{probe}(a, \theta_{probe}) = \sigma(\theta_{probe}^Ta)
    \label{eq:hpr_probe}
\end{equation}
where $\sigma(\cdot)$ denotes the sigmoid function and $\theta_{probe}$ is the weight vector of the probe. Readers may notice that Equation \ref{eq:hpr_probe} resembles a linear feedforward layer with no bias term. This is to ensure that the normal vector of the separating hyperplane passes through the origin, consistent with the direction-magnitude view.

At inference time, the probe weight vector is used to calculate a Householder matrix.
\begin{equation}
    H = I - \frac{2\theta_{probe}\theta_{probe}^T}{\theta_{probe}^T\theta_{probe}}
    \label{eq:probe_householder}
\end{equation}

The linear probe is trained using the Binary Cross Entropy (BCE) loss.
\begin{multline}
    % \centering
    \mathcal{L}_{probe} = \frac{1}{NS^y}\sum_{i=1}^N \sum_{j=1}^{S^y} \Bigl[ BCE(\sigma(\theta_{probe}^Ta_{i, j}^\mathbf{p}), 1) \\ + BCE(\sigma(\theta_{probe}^Ta_{i, j}^\mathbf{n}), 0) \Bigr]
    \label{eq:loss_probe}
\end{multline}

\subsection{Angle Prediction}
\label{sec:hpr_angle_pred}
Given the separating hyperplane for a layer, we seek to predict a rotating angle that helps transform the reflection of each negative activation into the desirable positive activation. As such, our key assumption considers the desirable positive activation vector to lie on the $2$-D plane formed by the original negative activation and its reflection, allowing us to efficiently perform the rotation of the negative activation vector. To this end, we employ a feedforward neural network $MLP$ to predict the rotating angle $f_{angle}(a, \theta_{angle})$ for an input vector $a$:
\begin{equation}
    \centering
    f_{angle}(a, \theta_{angle}) = \pi \times \sigma(MLP(a, \theta_{angle}))
    \label{eq:angle_pred}
\end{equation}
where $\theta_{angle}$ represents the model parameters. The output of $f_{angle}$ is a scalar value in the range $[0, \pi]$ radians.

Among several possible implementations, given a negative activation $a_{i,j}^\mathbf{n}$, we train $f_{angle}$ to predict the angle between the corresponding desired positive activation $a_{i,j}^\mathbf{p}$ and $a_{i,j}^\mathbf{n}$ for rotation. In contrast, if the input vector is a positive activation $a_{i,j}^\mathbf{p}$, $f_{angle}$ should return zero (i.e., no rotation). Our training loss for $f_{angle}$ is thus:
\begin{equation}
    \centering
    g(a_{i,j}^{\mathbf{p}}, a_{i,j}^{\mathbf{n}}) = \arccos(\frac{(a_{i,j}^{\mathbf{p}})^T a_{i,j}^\mathbf{n}}{\lVert a_{i,j}^\mathbf{p} \rVert \lVert a_{i,j}^\mathbf{n} \rVert})
    \label{eq:angle_func}
\end{equation}
\begin{multline}
    % \centering
    \mathcal{L}_{angle} = \frac{1}{NS^y}\sum_{i=1}^N \sum_{j=1}^{S^y} \biggl[ \Bigl( f_{angle}(a_{i, j}^\mathbf{n}, \theta_{angle}) \\ 
    - g(a_{i,j}^{\mathbf{p}}, a_{i,j}^{\mathbf{n}}) \Bigr)^2 \\ 
    + f_{angle}(a_{i, j}^\mathbf{p}, \theta_{angle})^2  \biggr]
    \label{eq:loss_angle}
\end{multline}
where $g(\cdot, \cdot)$ computes the angle between two vectors using the inverse of cosine $\arccos$. For training, the linear probe and angle prediction modules are optimized jointly via: $\mathcal{L} = \mathcal{L}_{probe} + \mathcal{L}_{angle}$.

\subsection{Computing the Final Activation}
\label{sec:hpr_compute_final}

At inference time, let $a$ be an activation in a layer of LLMs, we first forward it through the corresponding linear probe and the angle prediction module.
\begin{equation}
    \hat{\sigma} = \lfloor f_{probe}(a, \theta_{probe}) \rceil
    \label{eq:sign}
\end{equation}
\begin{equation}
    \gamma_1 = f_{angle}(a, \theta_{angle})
    \label{eq:gamma1}
\end{equation}
$\hat{\sigma}$ is rounded to the nearest integer, $0$ or $1$ to be specific, and predicts whether the given activation $a$ is positive or negative. If $a$ is predicted as a negative activation, we edit it by first reflecting $a$ about the separating hyperplane $\theta_{probe}$ to obtain the reflected vector $\dot{a}$ in the positive region. Afterward, we calculate a new activation by rotating $a$ within the $2$-D plan formed by $a$ and $\dot{a}$ by an angle of $\gamma_1$ radians. The resulting vector $\hat{a}$ will serve as our predicted positive activation for $a$.

%forms an angle of $\gamma_1$ radians with $a$ and

%we calculate a new activation vector $\hat{a}$ in the positive region such that it belongs to the $2$-D plan .

In particular, a Householder matrix is computed from the probe's weight following Equation \ref{eq:probe_householder}. With this we can reflect $a$ to obtain the reflected activation $\dot{a}$ and the angle $\gamma_2$ between $a$ and $\dot{a}$:
\begin{equation}
    \dot{a} = Ha, \text{  } \gamma_2 = g(\dot{a}, a)
    \label{eq:hat_a}
\end{equation}

\iffalse
\begin{equation}
    \dot{a} = Ha
    \label{eq:hat_a}
\end{equation}
\begin{equation}
    \gamma_2 = g(\dot{a}, a)
    \label{eq:gamma2}
\end{equation}
\fi

%Combining this with the fact that we have
\iffalse Since $a$ and $\dot{a}$ pass through the origin, there exists a $2$-D plane containing both of them. We assume $\hat{a}$ to also lie on this plane, making an approximation to the exact rotation. Moreover,\fi The Householder reflection and rotation transformation preserve vector norm. Thus, the norm of $a$, $\dot{a}$ and $\hat{a}$ are identical. Combined with the computed angles $\gamma_1$ and $\gamma_2$, the rotation on $2$-D plane \iffalse final activation $\hat{a}$ \fi to obtain the predicted positive activation $\hat{a}$ can be calculated via $a$ and $\dot{a}$ as follows:
\begin{equation}
    \centering
    \hat{a} = \frac{\sin(\gamma_1)}{\sin(\gamma_2)} \dot{a} + \frac{\sin(\gamma_2 - \gamma_1)}{\sin(\gamma_2)} a
    \label{eq:rotated_activation}
\end{equation}
The proof for Equation \ref{eq:rotated_activation} is in Appendix \ref{sec:appendix_proof}.

Finally, HPR's editing function can be written as follows: $f(a | \theta_{probe}, \theta_{angle}) = \hat{\sigma} a + (1-\hat{\sigma}) \hat{a}$.

\iffalse
\begin{equation}
    \centering
    f(a | \theta_{probe}, \theta_{angle}) = \hat{\sigma} a + (1-\hat{\sigma}) \hat{a}
    \label{eq:final_activation}
\end{equation}
This function essentially edits vectors identified as negative activation by the linear probe while retaining positive activation vectors.
\fi

\section{Experiments}
\label{sec:exp}

\subsection{Experimental Setup}
\label{sec:exp_setup}

\noindent \textbf{Datasets}: Following previous activation editing work \cite{NEURIPS2023_81b83900}, we first evaluate the models on the TruthfulQA dataset \citep{lin-etal-2022-truthfulqa}. TruthfulQA includes 817 questions, each of which is coupled with factually correct and incorrect answers. \iffalse The incorrect answers are based on many of human's false beliefs and misconceptions.\fi We split the dataset into subsets with ratios $45$ / $5$ / $50$ for training, validation and testing respectively.

Aside from truthfulness, we also demonstrate the proposed method on other societal issues related to LLMs, more specifically, bias, ethics, and toxicity. These are reflected in BigBench's Bias Benchmark for QA (BBQ) (\citealp[]{srivastava2023beyond}; \citealp{parrish-etal-2022-bbq}), BigBench's Simple Ethical Questions (SEQ), and Toxigen \citep{hartvigsen-etal-2022-toxigen}, respectively. 
These datasets are already split into a training set and a validation set. We use the validation sets to test the models, while splitting their training sets further with ratios $90$ / $10$ to make new training and validation sets.

All four datasets are multiple choice tasks, thus the main evaluation metrics is multiple choice accuracy. The correct and incorrect answers for each question can be used handily to create $y_i^\mathbf{p}$ / $y_i^\mathbf{n}$ pairs.

%The only exception is BBQ, which we randomly sample $6000$ examples from its training set for training and validation.

\noindent \textbf{Base Models and Baselines}: We conduct experiments with three recent popular open source LLMs: \texttt{Llama2-7B-Chat} \citep{touvron2023llama2}, \texttt{Mistral-7B-Instruct} \citep{jiang2023mistral}, and \texttt{Llama3-8B-Instruct} \citep{llama3modelcard}. We compare our method with the following baselines: 

%These are LLMs of roughly the same size but with varied reasoning capability.

%\noindent \textbf{Baselines}: 

%\begin{itemize}
    %\item 
    $\bullet$ \textbf{Base}: The unaltered base LLMs.
    
    %The selected models are listed above.
    
    %\item 
    $\bullet$ \textbf{LoRA} \citep{hu2022lora}: We fine-tune the base LLM with LoRA adapter on the same training data as activation editing methods for a fair comparison.

    %($r=8$, $\alpha=8$)
    %\item 
    $\bullet$ \textbf{Diff}: Given a positive or negative activation $a_{i,j}$, this baseline employs a feedforward network to directly predict the difference vector $a_{i,j}^{\mathbf{p}}-a_{i,j}$ with the corresponding positive activation $a_{i,j}^{\mathbf{p}}$. At inference time, we utilize the sum of the original activation vector $a_{i,j}$ and its predicted difference vector to obtain the predicted positive activation.
    
    %\item 
    $\bullet$ \textbf{ITI} \citep{NEURIPS2023_81b83900}: A representative Activation Editing method for the aforementioned points-in-space view that shifts the outputs of a set of attention heads in each layer by a fixed steering direction. The steering vector in ITI is the Mass Mean Shift vector (i.e. the difference between the centers of the positive and negative regions) of activations in training data (i.e., not learnable). We employ the source code published by the original authors. However, their code is implemented only for Llama models and TruthfulQA dataset specifically. Thus we only report results of ITI with \texttt{Llama2-7B-Chat} and \texttt{Llama3-8B-Instruct} on TruthfulQA.
    
    %A representative Activation Editing method that shifts the outputs of a set of attention heads in each layer by a fixed steering direction. This work is representative of the aforementioned points-in-space view. We employ the source code published by the original authors. However, their code is implemented only for Llama models and TruthfulQA dataset specifically. Thus we only report results of ITI with \texttt{Llama2-7B-Chat} and \texttt{Llama3-8B-Instruct} on TruthfulQA.
%\end{itemize}

\noindent \textbf{Evaluation Framework}: We utilize EleutherAI's Language Model Evaluation Harness \citep{eval-harness}, a reliable evaluation framework used in numerous works including HuggingFace's Open LLM Leaderboard. This framework supports automatic evaluation of various benchmark datasets for LLM. Our experiments involve evaluating mulitple choice accuracy on various datasets. This is done by calculating the aggregated log-likelihood of each choice given the input prompt and then pick the top one.

%Our linear probe and angle prediction modules do not have many hyperparameters.

\noindent \textbf{Hyperparameters}: In our model, the linear probe is a vector of the same dimensions as the LLMs' hidden dimensions. The angle prediction module is a feedforward neural network with 4 layers and one output unit. We train each module for 5 epochs with batch size $16$, AdamW optimizer \citep{loshchilov2018decoupled}, learning rate $5 \times 10^{-4}$, cosine learning rate scheduler and warmup. For editing, we apply HPR to the top $k=5$ layers with the highest probe accuracy. Appendix \ref{sec:appendix_num_layers} presents model performance with different values of $k$. % We also provide a reproducibility checklist in Appendix \ref{app:repo}.

%over the first $5\%$ of the training duration

%as described in Section \ref{sec:hpr_angle_pred}

%We examine the accuracy of linear probe in each layer and find that middle layers of the LLM tend to separate activations the best, as can be seen in Figure \ref{fig:probe_acc_llama2}. This observation aligns with previous work (\citealp[]{NEURIPS2023_81b83900}; \citealp[]{joshi2024personas}) suggesting that editing these layers may yield the best results. For evaluation, we choose the top $k$ layers with the highest probe accuracy to edit, and tune $k$'s value from $1$ to $5$.

\subsection{Results}
\label{sec:res_tqa}

% Please add the following required packages to your document preamble:
% \usepackage{multirow}
\begin{table}[h]
\centering
\resizebox{\columnwidth}{!}{%
\begin{tabular}{l|cccccc}
\hline
\multicolumn{1}{l|}{\multirow{3}{*}{\textbf{Method}}} & \multicolumn{6}{c}{\textbf{Model}}                                                                               \\\cline{2-7}
\multicolumn{1}{c|}{}                                 & \multicolumn{2}{c}{\textbf{Llama2}} & \multicolumn{2}{c}{\textbf{Llama3}} & \multicolumn{2}{c}{\textbf{Mistral}} \\
\multicolumn{1}{c|}{}                                 & \textbf{MC1}     & \textbf{MC2}     & \textbf{MC1}     & \textbf{MC2}     & \textbf{MC1}      & \textbf{MC2}     \\ \hline
\multirow{2}{*}{Base}                                 & 29.58            & 43.00            & 36.43            & 50.73            & 54.28             & 67.45            \\
                                                      & ± 2.26           & ± 2.17           & ± 2.38           & ± 2.13           & ± 2.47            & ± 2.14           \\ \cline{2-7}
\multirow{2}{*}{LoRA}                                 & 29.10            & 43.40            & 38.63            & 55.84            & 54.77             & 70.45            \\
                                                      & ± 2.25           & ± 2.15           & ± 2.41           & ± 2.11           & ± 2.46            & ± 2.06           \\ \cline{2-7}
\multirow{2}{*}{Diff}                                 & 33.74            & 48.87            & 29.34            & 52.53            & 50.61             & 68.68            \\
                                                      & ± 2.47           & ± 2.24           & ± 2.25             & ± 2.25             & ± 2.48            & ± 2.11           \\ \cline{2-7}
\multirow{2}{*}{ITI}                                  & 33.74            & 50.67            & 39.85            & 56.58            & -                 & -                \\
                                                      & ± 2.34           & ± 2.20           & ± 2.42           & ± 2.18           & -                 & -                \\ \hline
\multirow{2}{*}{{\bf HPR}}                           & \textbf{51.83}   & \textbf{70.95}   & \textbf{52.32}   & \textbf{71.70}   & \textbf{55.01}    & \textbf{72.14}   \\
                                                      & ± 2.47           & ± 2.12           & ± 2.47           & ± 2.13           & ± 2.46            & ± 2.07           \\ \cline{2-7}
\multirow{2}{*}{-AnglePred}                         & 30.07            & 43.36            & 35.94            & 49.77            & 53.79             & 67.31            \\
                                                      & ± 2.27           & ± 2.18           & ± 2.375          & ± 2.12           & ± 2.47            & ± 2.14           \\ \hline
\end{tabular}%
}
\caption{Model performance (in $\%$) on TruthfulQA multiple choice tasks. ± indicates standard errors.}
\label{table:res_tqa}
\end{table}

% Please add the following required packages to your document preamble:
% \usepackage{multirow}
\begin{table}[h]
\centering
\resizebox{0.9\columnwidth}{!}{%
\begin{tabular}{l|ccc}
\hline
\multicolumn{1}{c|}{\multirow{2}{*}{\textbf{Model}}} & \multicolumn{3}{c}{\textbf{Dataset}}                                   \\ \cline{2-4} 
\multicolumn{1}{c|}{}                                & \textbf{BBQ}   & \textbf{SEQ}                       & \textbf{Toxigen} \\ \hline
Llama2-7B-Chat                                       & 33.27          & 21.74                              & 51.38            \\
\multicolumn{1}{r|}{+ HPR}                           & \textbf{38.38} & \textbf{60.87}                     & \textbf{52.34}   \\ \hline
Llama3-8B-Instruct                                   & 60.44          & 47.83                              & 45.32            \\
\multicolumn{1}{r|}{+ HPR}                           & \textbf{67.10}  & \textbf{52.17}                     & \textbf{46.81}   \\ \hline
Mistral-7B-Instruct                                  & 61.62          & 69.57                              & 55.00            \\
\multicolumn{1}{r|}{+ HPR}                           & \textbf{73.24} & \multicolumn{1}{r}{\textbf{86.96}} & \textbf{61.60}   \\ \hline
\end{tabular}%
}
\caption{HPR performance for bias, ethics, and toxicity. We report multiple choice accuracy in $\%$.}
\label{table:other_tasks}
\end{table}

\noindent \textbf{TruthfulQA}: Table \ref{table:res_tqa} presents the performance of our method HPR and the baselines on TruthfulQA. The results include both MC1, multiple choices with only 1 correct answer per question, and MC2, which is multiple choices with more than 1 correct answer for each question. The first observation from the table is that fine-tuning LLMs with LoRA does not produce consistent performance improvement for TruthfulQA over different models. In contrast, activation editing methods, i.e., ITI and HPR, consistently outperform the base LLM models, achieving greater margins than LoRA fine-tuning. It thus highlights the effectiveness of activation editing for altering LLMs for desirable behaviors. When comparing Diff and ITI, ITI's superior overall performance indicates that learning negative-positive difference vectors for activations, as done in Diff, is ineffective and cannot ensure optimal aligning performance for LLMs. Most importantly, the proposed model HPR is significantly better than all the baselines with substantial margins across all base LLMs. These results clearly testify to the advantages of HPR, demonstrating the benefits of our new direction-magnitude view for activation editing with reflection and rotation for negative activation transformation.

\noindent \textbf{Ablation Study}: The last row in Table \ref{table:res_tqa} further shows the performance of HPR when the angle prediction module is excluded from the full model. In other words, the editing function now only reflects negative activation vectors about the separating hyperplane defined by the linear probe. As can be seen, this exclusion leads to significant performance drops across all base LLMs for HPR, suggesting that simply having the activations landed on the positive region is not enough to make an effective edit. Thereby, it justifies the importance of angle prediction to adjust reflected activations for our model. We also note that the linear probe module cannot be removed from HPR for ablation study as it is essential for finding the positive-negative separating hyperplane and rotating plane in our model. Finally, the superior performance of HPR for different LLMs confirms the advantages of our assumption on the shared $2$-D plane of $a$, $\dot{a}$, and $\hat{a}$.

%for the original negative activations, the reflected activations, and the desirable positive activations.

% including the activation editing method ITI,

%It is clear from the table that HPR significantly outperforms all baselines by significant margins. The difference between HPR and the baselines is higher for Llama models while this number is lower in the case of Mistral. This may be due to Activation Editing not introducing any new knowledge but only altering the behavior of the LLM. 

%Surprisingly, LoRA fine-tuned models, which is supposed to be the upper-bound in terms of performance, is actually not much better than their respective base model. The explanation for this can be the small number of training examples. It proves that Activation Editing methods are a better choice than fine-tuning when training data is scarce.

\begin{figure*}[hbt!]
    \centering
    \begin{subfigure}{0.7\textwidth}
        \includegraphics[width=\linewidth]{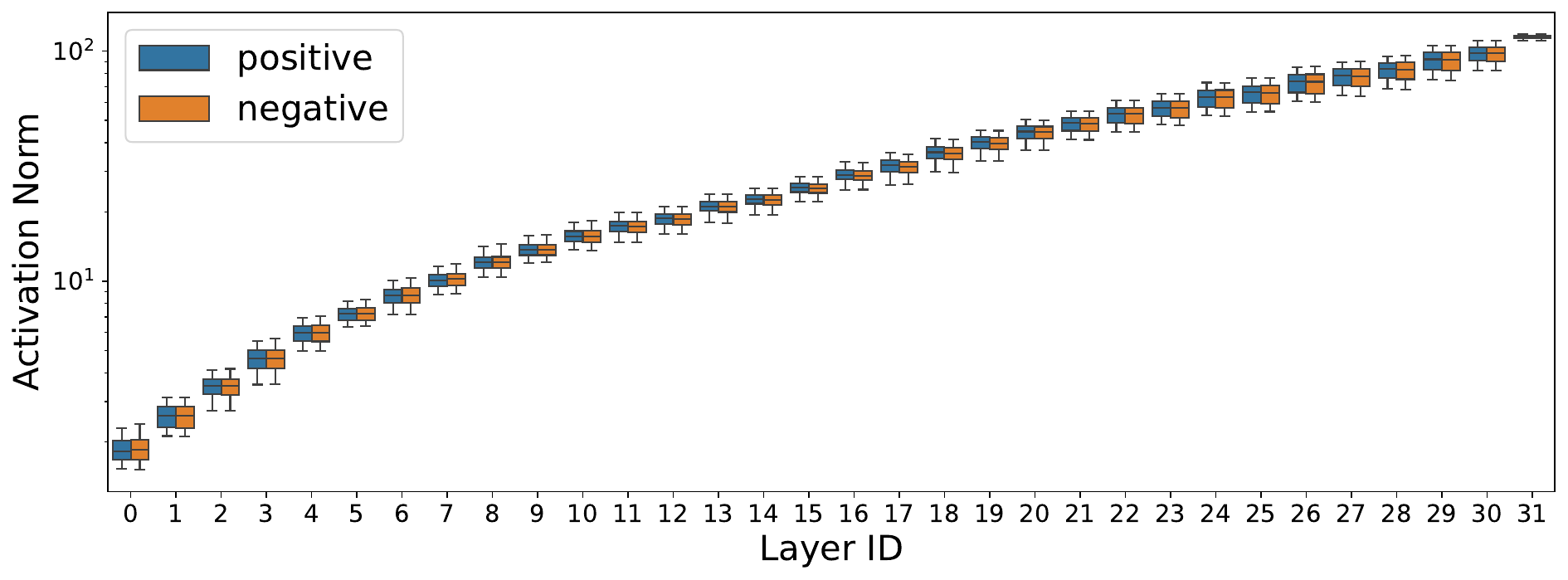}
        \caption {\texttt{Llama2-7B-Chat}}
        \label{fig:norm_dist_llama2}
    \end{subfigure}
    \begin{subfigure}{0.7\textwidth}
        \includegraphics[width=\linewidth]{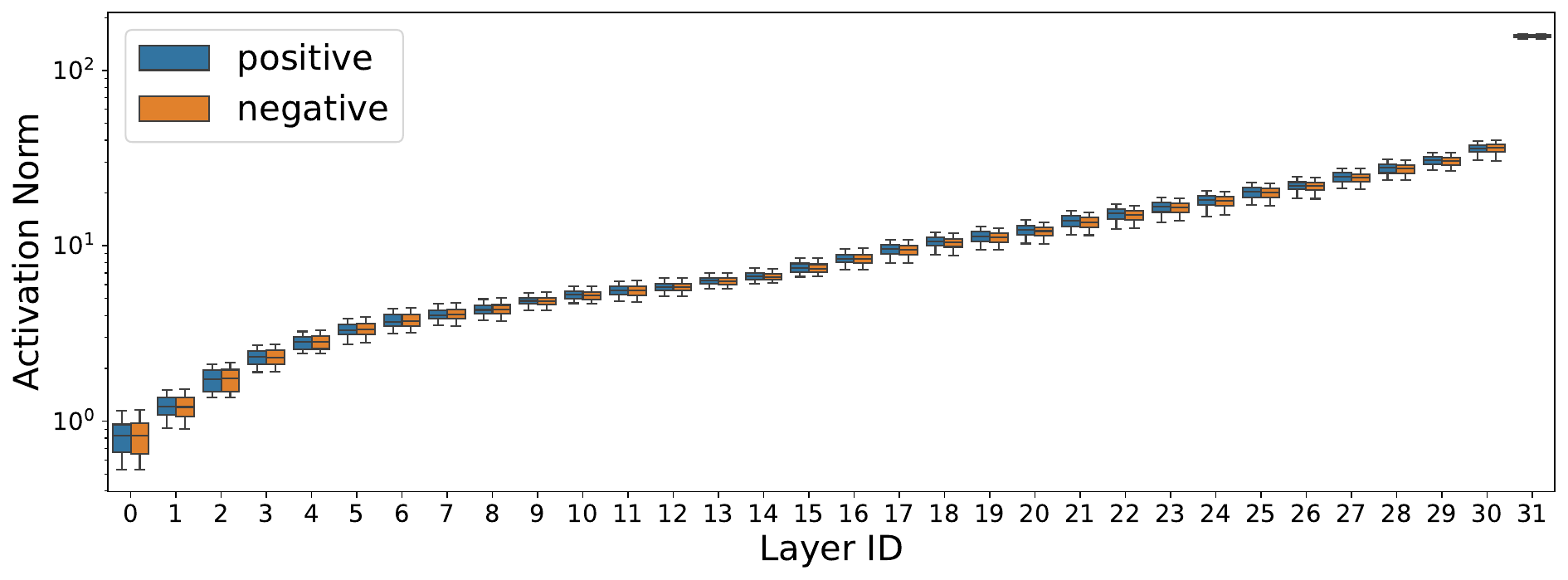}
        \caption {\texttt{Llama3-8B-Instruct}}
        \label{fig:norm_dist_llama3}
    \end{subfigure}
    \begin{subfigure}{0.7\textwidth}
        \includegraphics[width=\linewidth]{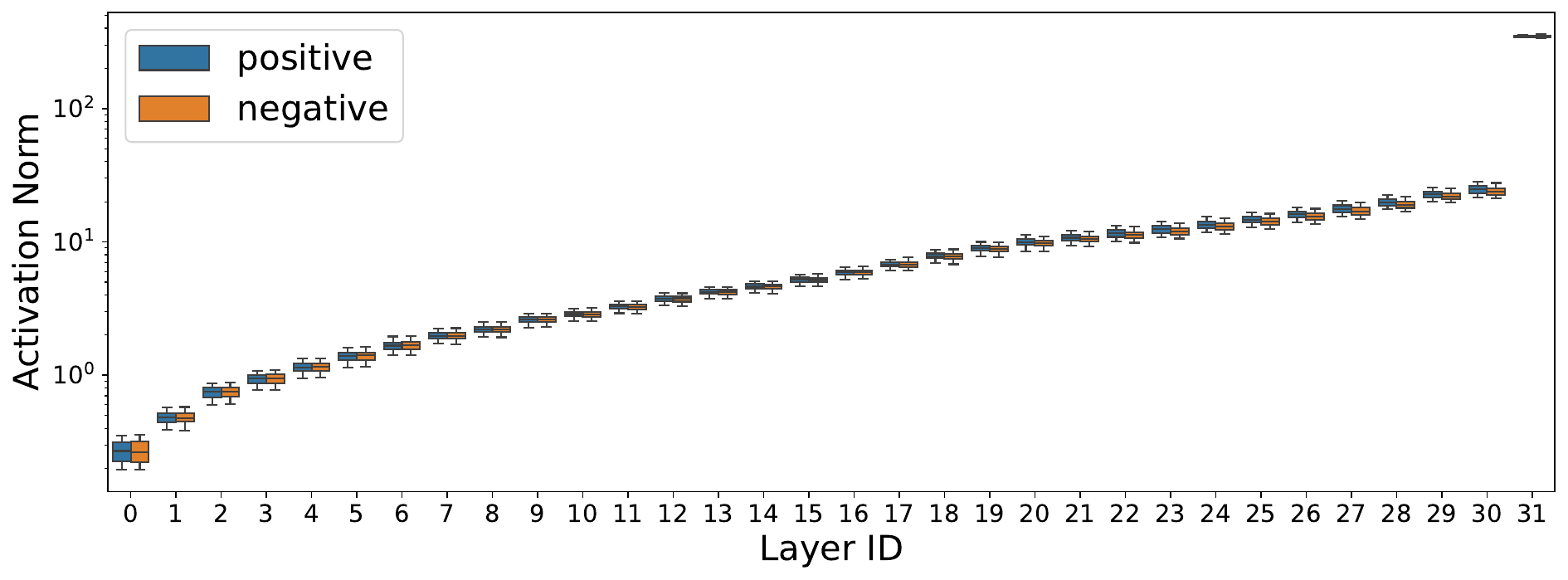}
        \caption {\texttt{Mistral-7B-Instruct}}
        \label{fig:norm_dist_mistral}
    \end{subfigure}

    \caption{The activation norms in $log_{10}$ scale across $32$ transformer blocks of three popular LLMs. Each box plot represents the norm distribution in a layer of the LLMs.}
    % We present the norm distribution in each layer as box plots, each box represents a layer. The incremental medians show that activation norms tends to increase as the layer becomes deeper, while the small vertical length of these boxes suggests that the variation in vector norm of activations of the same layer is extremely low.
    \label{fig:norm_dist}
\end{figure*}

\begin{figure*}[hbt!]
    \begin{subfigure}{0.33\textwidth}
        \includegraphics[width=\linewidth]{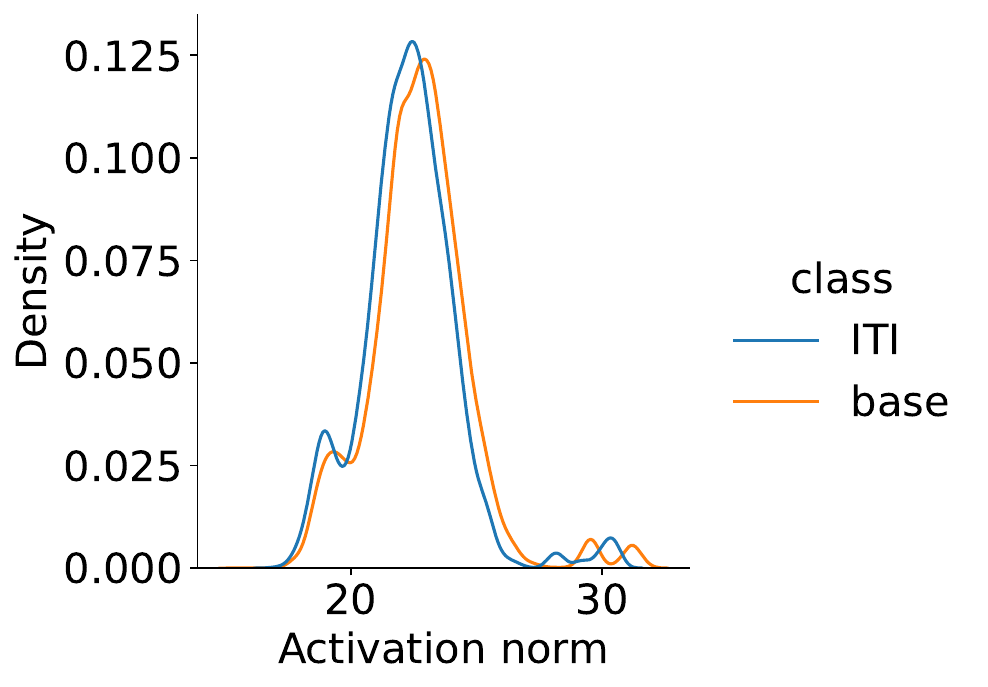}
        \caption {ITI, $\alpha=15$}
        \label{fig:act_norm_dist_iti_15}
    \end{subfigure}
    \begin{subfigure}{0.33\textwidth}
        \includegraphics[width=\linewidth]{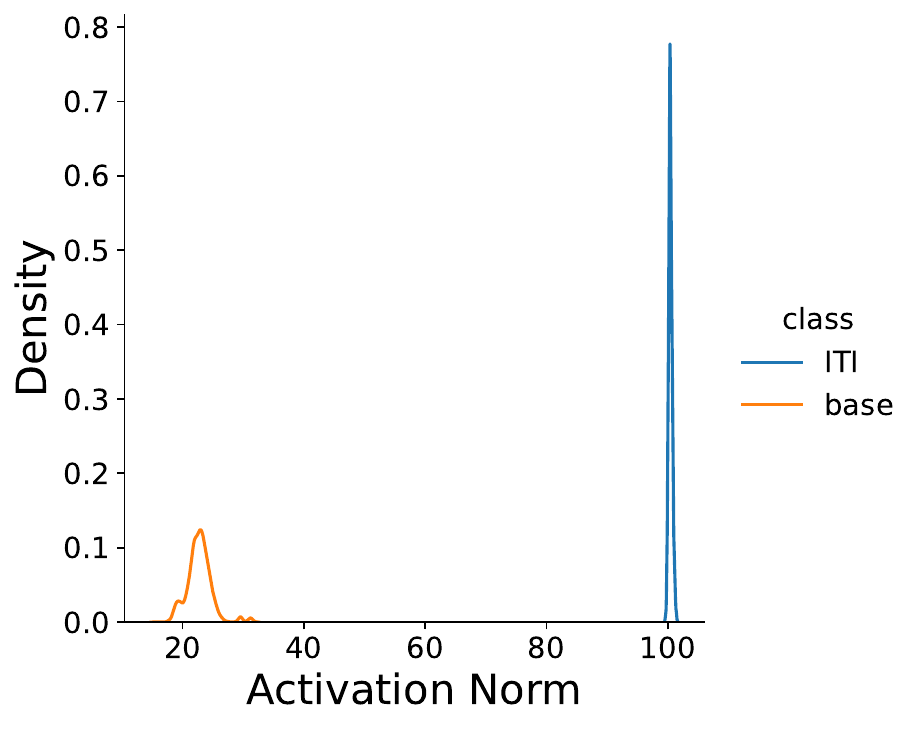}
        \caption {ITI, $\alpha=200$}
        \label{fig:act_norm_dist_iti_50}
    \end{subfigure}
    \begin{subfigure}{0.33\textwidth}
        \includegraphics[width=\linewidth]{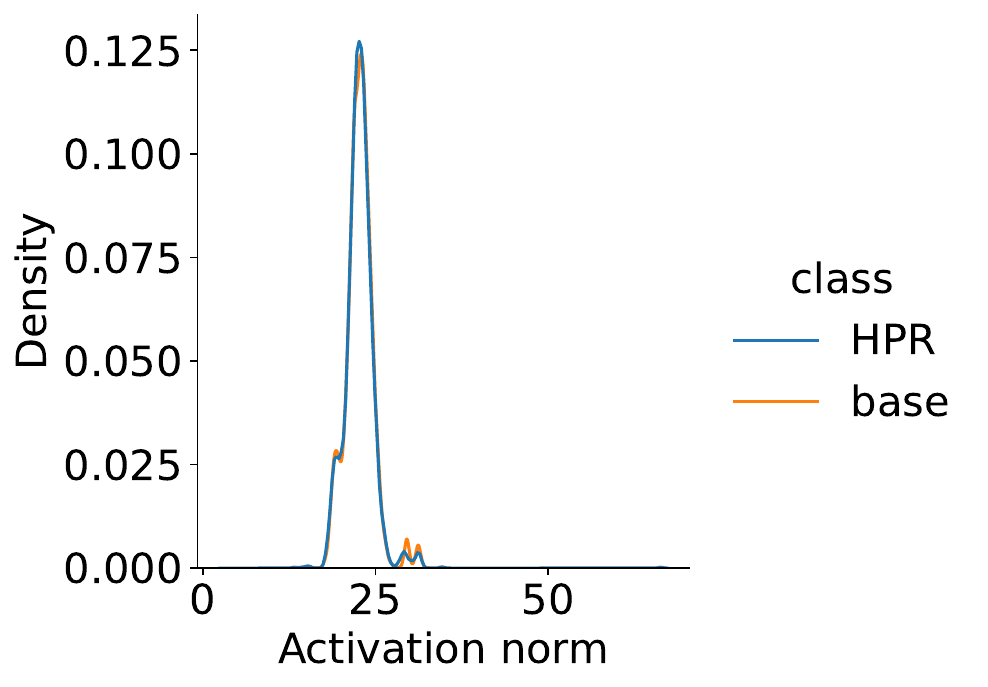}
        \caption {HPR}
        \label{fig:act_norm_dist_hpr}
    \end{subfigure}

    \caption{Activation norm distributions of the $14^{th}$ layer of \texttt{Llama2} before and after being edited. We use the $14^{th}$ layer as it has the highest probe accuracy in Figure \ref{fig:probe_acc_llama2}. Similar trends can be seen for other layers and models.}
    \label{fig:act_norm_dist}
\end{figure*}

% \textcolor{red}{``tasks''?? Perhaps we should change this word into something else}

%\label{sec:other_tasks}

\noindent \textbf{BBQ, SEQ, and Toxigen}: To further illustrate the effectiveness of HPR in eliciting desirable behavior, Table \ref{table:other_tasks} shows HPR's performance on the BBQ, SEQ, and Toxigen datasets. These datasets evaluate the abilities of LLMs to generate unbiased (BBQ), ethically acceptable (SEQ), and non-toxic (Toxigen) responses. Across various base LLMs, incorporating HPR can significantly enhance performance on all of these datasets. These results highlight the benefits of HPR in improving important safety criteria for LLMs, leading to unbiased, ethical, and non-toxic responses for responsible models.

\subsection{Analysis of Activation Space}
\label{sec:analysis_act}

In this section, we examine the activation norms of the selected LLMs to gain a better understanding of the activation space. We first look into base LLMs. In Figure \ref{fig:norm_dist} we plot the activation norms in each layer, positive vectors and negative vectors side-by-side. From these box plots, we can observe the \textbf{Magnitude Consistency} property: activations of the same layer have roughly the same vector norm for all considered LLMs. This observation holds true regardless of the activations being positive or negative. The gap between the whiskers of each box is very narrow, suggesting a low variance. This gap seems to become narrower for more powerful models, as can be seen in Figures \ref{fig:norm_dist_llama3}, \ref{fig:norm_dist_mistral} for LLaMA3 and Mistral. Due to this universality, we consider activation norm consistency as a necessary condition that should be maintained by editing methods to achieve desired LLMs.

Considering this property, we demonstrate how the steering vector approach in ITI \cite{NEURIPS2023_81b83900} struggles to simultaneously maintain activation magnitude consistency and effectively alter their activations for greater improvement on desired behaviors. First, Figures \ref{fig:act_norm_dist_iti_15} and \ref{fig:act_norm_dist_iti_50} show the distributions of activation norms in LLMs before and after editing with ITI. In Figure \ref{fig:act_norm_dist_iti_15}, the scaling factor $\alpha$ is set to $15$ (i.e., ITI$_{15}$), as recommended in the original ITI paper, while in Figure \ref{fig:act_norm_dist_iti_50}, $\alpha$ is set to $200$ (i.e., ITI$_{200}$). As can be seen, the smaller scaling factor $\alpha=15$ in ITI$_{15}$ leads to less divergence of activation norms than ITI$_{200}$ from the original LLMs (i.e., better preservation of activation norms).

What is the implication of such slight norm divergence from base LLMs for ITI? In Table \ref{table:behavior_change}, we present the behavior shift rates of ITI$_{15}$ and ITI$_{200}$ compared to the original \texttt{Llama2-7B-Chat} model on TruthfulQA. Specifically, we show how often each editing method can flip the LLM's predictions of examples from true to false and vice versa. From the table, we observe that the slight divergence of activation norms in ITI$_{15}$ results in a more limited ability to change the base model's behavior, with a behavior shift rate of only $8.56\%$ compared to $34.23\%$ for ITI$_{200}$. As the behavior shift rate is the upper bound of the overall performance improvement on TruthfulQA for ITI, this limited ability to alter LLM behavior will hinder further improvement with a small scaling factor in ITI.

% Please add the following required packages to your document preamble:
% \usepackage{multirow}
\begin{table}[h]
\centering
\resizebox{\columnwidth}{!}{%
\begin{tabular}{l|cccc|c}
\hline
\multicolumn{1}{c|}{\multirow{2}{*}{\textbf{Model}}} & \textbf{False to}        & \textbf{True to}              & \textbf{Remains}          & \textbf{Remains}              & \textbf{Overall}            \\
\multicolumn{1}{c|}{}                                & \textbf{True}$\uparrow$  & \textbf{False}$\downarrow$    & \textbf{True}$\uparrow$   & \textbf{False}$\downarrow$    & \textbf{Acc.}$\uparrow$     \\ \hline
Base model                                           & -                        & -                             & 29.58                     & 70.42                         & 29.58                       \\ \hline
ITI, $\alpha=15$                                     & 6.36                     & \textbf{2.20}                 & \textbf{27.38}            & 64.06                         & 33.74                       \\
% ITI, $\alpha=50$                                     & 11.25                    & 8.31                          & 21.27                     & 59.17                         & 32.52                       \\
ITI, $\alpha=200$                                    & 14.18                    & 20.05                         & 9.54                      & 56.23                         & 23.72                       \\
HPR                                                  & \textbf{28.85}           & 6.60                          & 22.98                     & \textbf{41.56}                & \textbf{51.83}              \\ \hline
\end{tabular}%
}
\caption{Behavior shift rate (in $\%$) of activation editing methods on TruthfulQA MC1 task compared to the base model. The base LLM is \texttt{Llama2-7B-Chat}. $\uparrow$ means greater is better and $\downarrow$ means lower is better.}
\label{table:behavior_change}
\end{table}

Furthermore, with a larger scaling factor of $\alpha=200$, the greater behavior shift rate in ITI$_{200}$ might suggest that ITI$_{200}$ can better boost truthful performance for ITI. However, a closer examination at Table \ref{table:behavior_change} reveals that the significant norm change in ITI$_{200}$ promotes both ``good'' False-to-True and ``bad'' True-to-False prediction flips from the base LLM. While  ITI$_{200}$ is more effective at correcting false predictions, increasing the ``False-to-True'' flip rate from $6.36\%$ in ITI$_{15}$ to $14.18\%$, it also introduces more ``bad'' edits, changing $20.05\%$ of examples with True predictions in the base LLM to False, compared to just $2.2\%$ for ITI$_{15}$. Overall, the bad edits significantly dominate the good edits in the ITI model with more extensive norm change, ITI$_{200}$, leading to its poorer performance in producing truthful responses. To this end, our analysis demonstrates the fundamental limitations of steering vector approach on boosting truthful performance for LLMs, regardless of efforts to tune the scaling factor.

In contrast, Figure \ref{fig:act_norm_dist_hpr} highlights the inherent ability of the proposed HPR method to preserve activation norms through its activation rotation mechanisms. In addition, HPR offers substantially stronger editing capabilities for achieving desired behaviors in LLMs as shown in Table \ref{table:behavior_change}. It significantly improves the False-to-True prediction flip rate while minimizing undesirable True-to-False edits for the base LLM, demonstrating the effectiveness of our method for activation editing.

\subsection{Impact on Generation Capability}
\label{sec:generation}

% TODO: HERE
\textbf{Generation Quality}: To assess the impact of the proposed method on generation capability, we perform evaluations in the open-ended generation setting of TruthfulQA with LLaMA models. For this evaluation, we employ BLEU accuracy, which is calculated as the ratio of generated responses having BLEU scores with their respective correct (positive) references higher than that with the incorrect (negative) references, as described in \citet{lin-etal-2022-truthfulqa}. We use the popular implementation in \citet{eval-harness} and the same data split as in Section \ref{sec:exp_setup}. The results are presented in Table \ref{table:tqa_gen_bleu}, where ITI$_{15}$ and ITI$_{50}$ refer to the ITI method with a scaling factor $\alpha=15$ and $\alpha=50$ respectively.
\begin{table}[h]
\centering
\resizebox{0.85\columnwidth}{!}{%
\begin{tabular}{llc}
\hline
\multicolumn{1}{c}{\textbf{Backbone}} & \textbf{Method} & \textbf{BLEU Acc} \\ \hline
\multirow{4}{*}{Llama2-7B-Chat}       & Base            & 38.39             \\ 
                                      & ITI$_{15}$      & 41.56             \\  
                                      & ITI$_{50}$      & 34.96             \\ 
                                      & HPR             & \textbf{42.30}    \\ \hline
\multirow{4}{*}{Llama3-8B-Instruct}   & Base            & 43.28             \\  
                                      & ITI$_{15}$      & 41.32             \\ 
                                      & ITI$_{50}$      & 41.08             \\ 
                                      & HPR             & \textbf{44.74}    \\ \hline
\end{tabular}%
}
\caption{Automated evaluation of TruthfulQA open-ended generation task.}
\label{table:tqa_gen_bleu}
\end{table}

% TODO: HERE
% For human evaluation, we randomly sample 100 examples from the dataset and hire a native speaker to assess the responses from each method. They are instructed to judge whether the generated responses were truthful or not. The models' accuracy in producing truthful responses is shown in Table \ref{table:tqa_gen_human}.
% \begin{table}[h]
% \centering
% \begin{tabular}{l|ccc}
% \hline
%                    & \textbf{Base} & \textbf{ITI}_{15} & \textbf{HPR}          \\ \hline
% Llama2-7B-Chat     & 49            & 55                & \textbf{60}           \\
% Llama3-8B-Instruct & 56            & 55                & \textbf{63}           \\ \hline
% \end{tabular}
% \caption{Human evaluation result.}
% \label{table:tqa_gen_human}
% \end{table}

% TODO: HERE
It is clear from the table that the proposed HPR method also outperforms different baselines significantly on generation-based evaluation for TruthfulQA.

% TODO: HERE
In addition, we evaluate the fluency of the models' generated responses. Table \ref{table:fluency} shows the perplexity scores and bits per byte (smaller is better) on the test set of WikiText-2 \citep{merity2017pointer}.
\begin{table}[h]
\centering
\resizebox{\columnwidth}{!}{
\begin{tabular}{ll|ccc}
\hline
\multicolumn{1}{c}{\textbf{Backbone}} & \multicolumn{1}{c|}{\textbf{Method}} & \textbf{Word Ppl$\downarrow$} & \textbf{Byte Ppl$\downarrow$} & \textbf{BpB$\downarrow$} \\ \hline
\multirow{4}{*}{Llama2}               & Base                                 & 13.7077                       & 1.6316                        & 0.7063                             \\ \cline{2-5} 
                                      & ITI$_{15}$                           & 14.2038                       & 1.6425                        & 0.7158                             \\
                                      & ITI$_{50}$                           & 133.7374                      & 2.4982                        & 1.3209                             \\
                                      & HPR                                  & \textbf{13.7206}              & \textbf{1.6319}               & \textbf{0.7066}                    \\ \hline
\multirow{4}{*}{Llama3}               & Base                                 & 11.9524                       & 1.5903                        & 0.6693                             \\ \cline{2-5} 
                                      & ITI$_{15}$                           & 17.0515                       & 1.6996                        & 0.7652                             \\
                                      & ITI$_{50}$                           & 4303.4616                     & 4.7813                        & 2.2574                             \\
                                      & HPR                                  & \textbf{11.9558}              & \textbf{1.5904}               & \textbf{0.6694}                    \\ \hline
\end{tabular}%
}
\caption{Perplexity and bits-per-byte on WikiText-2 test set.}
\label{table:fluency}
\end{table}

% TODO: HERE
As can be seen, our method HPR does not need to sacrifice the models’ fluency to achieve effective editing for desirable model behavior, unlike the Steering Vector methods such as ITI. Notably, when the scaling factor of ITI is high (i.e., $50$), the perplexity scores become extremely large, leading to gibberish responses.

% \subsection{Impact on Inference Speed}
% \label{sec:inference_speed}

\begin{table}[h]
\centering
\resizebox{0.9\columnwidth}{!}{
\begin{tabular}{lccc}
\hline
           & \textbf{Llama2-7B} & \textbf{Llama3-8B} & \textbf{Mistral-7B} \\ \hline
Base       & 2.33               & 2.08               & 2.33                \\
LoRA       & 1.30               & 1.24               & 1.23                \\
ITI        & 2.06               & 1.87               & -                   \\
HPR$_{1}$  & 2.18               & 1.99               & 2.25                \\
HPR$_{5}$  & 1.98               & 1.80               & 1.98                \\
HPR$_{10}$ & 1.73               & 1.63               & 1.76                \\ \hline
\end{tabular}
}
\caption{Inference speed in samples per second (larger is better).}
\label{table:inference_speed}
\end{table}

% TODO: HERE
\noindent \textbf{Inference Speed}: Table \ref{table:inference_speed} compares HPR with the base model, ITI, and LoRA adapter in terms of inference speed. For HPR, we report the speed for its three variants corresponding to the number of edited layers. We see that with only one edited layer, the inference speed is slower than the base model but faster than LoRA and ITI. HPR slows down when we choose more layers to edit, which is natural. Importantly, we do not see significant speed reduction due to the introduction of our activation editing method, thus suggesting its potential applications in different scenarios.

% TODO: NEED REVISION
In fact, efficiency is a key motivation for the design of our method. Our method does not perform direct rotation for each activation in the models as it will be very expensive. Instead, we find a common hyperplane to separate the negative and positive regions of the activations, and we use this hyperplane to efficiently find the rotating direction for all activations in the layer (i.e., via the reflections). As finding rotating directions is the most expensive part for a rotation operation, using a common hyperplane for all activations significantly reduces our computation costs for editing. Finally, to compute the rotating angles, our method employs a regression model, which is very efficient and can be applied for each activation to improve the performance of our method.

\section{Related Work}
\label{sec:related_work}

%\noindent \textbf{Parameter-Efficient Fine-Tuning}: In the era of Large Language Models, where every model has billions of parameters, efficient tuning methods have become more crucial than ever. Prompt-Tuning \citep{lester-etal-2021-power} and Prefix-Tuning \citep{li-liang-2021-prefix} directly optimized continuous embedding tokens, which are prepended to the inpnut, in order to take advantage of the emergent ability and auto-regressive nature of language models. \citet{pmlr-v97-houlsby19a} introduced the idea of adding adapter modules to each Transformer layer. LoRA \citep{hu2022lora} expanded on this idea and became the dominant choice for fine-tuning LLMs. Our proposed method also train lightweight modules for each edited layer. However, we train them to guide the inner activations of the models instead of following the usual next token prediction objective.

%adapters \cite{pmlr-v97-houlsby19a}

Concerning the societal risks of LLMs, various approaches have been explored to control and align their behavior post-pretraining. Unlike resource-intensive methods for LLM alignment such as instruction tuning and reinforcement learning from human feedback \cite{NEURIPS2022_b1efde53,bai2022training}, our work falls into the category of resource-efficient methods for controlling LLMs. Several resource-efficient approaches exist in this area. First, parameter-efficient fine-tuning aims to fine-tune LLMs with safety data while minimizing the number of learnable parameters, such as prompt-tuning \citep{lester-etal-2021-power}, prefix-tuning \citep{li-liang-2021-prefix}, and LoRA \citep{hu2022lora}. However, fine-tuning might also compromise the safety of LLMs \cite{qi2023finetuning}. Additionally, model editing attempts to locate and edit model parameters associated with safety issues using minimal invasions for efficiency \cite{meng2022locating,ilharco2023editing}. However, model editing might impact the general robustness of the models \cite{brown2023robustness}. Our work belongs to the third direction for efficient LLM control, i.e., activation editing, which involves editing their inner representations towards a desired behavior at inference time \cite{li2023emergent,hernandez2023remedi} and can be traced back to plug-and-play controllable text generation research \cite{Dathathri2020Plug,krause-etal-2021-gedi-generative}. Accordingly, activation editing can preserve the pretrained LLMs to achieve better robustness while still offering adjustable and minimally invasive benefits.

In one approach to activation editing, \citet{liu-etal-2021-dexperts}, \citet{li-etal-2023-contrastive}, and \citet{liu2024tuning} contrast the behavior of an expert and an amateur model. Additionally, vector steering edits inner representations by adding steering vectors (\citealp[]{burns2023discovering}; \citealp[]{NEURIPS2023_81b83900}; \citealp[]{turner2023activation}; \citealp[]{rimsky2024steering}; \citealp[]{vonrütte2024language}). However, none of these work explores the direction-magnitude perspective with activation rotations.

\section{Conclusion}
\label{sec:conclusion}

This work proposes a new activation editing approach based on the direction-maginitude view. By rotating negative activations instead of adding to them a fixed steering vector, our proposed method effectively addresses the shortcomings of existing work, as evidenced by the improved performance on various benchmarks. Our analyses highlight the magnitude consistency property of LLMs, providing insights into the operations of our editing method. In the future, we plan to extend our research to study how the activation space evolves during fine-tuning and how the proposed method scales to larger models and other architectures. 

%This observation sheds some light on the current understanding of LLMs' inner representation. Our paper can provide useful insights for future work as well as leave open questions to be answered, such as how the activation space evolves during fine-tuning, how different pre-training strategies impact the activation space, how to adapt HPR to larger models, models of different architectures or with mixture of experts, etc.

\section*{Limitations}
\label{sec:limitations}
As an empirical study, our work is not without limitations. Acknowledging this, we would like to discuss them as follows:
\begin{itemize}
    \item \iffalse \textbf{Model size}: Our work is a proof of concept for HPR editing method. \fi Due to limited computational resources, we only employ open-source LLMs of size $7$-$8$B parameters. However, we show that the proposed method can effectively alter the behaviors of LLMs for diverse base models and tasks. We leave further research on the scalability of HPR as well as its impact on models of larger sizes for future work.
    
    \item Although our method exhibits strong editing performance for desired behaviors, the method itself, like all other Activation Editing methods, only serves to alter LLMs' behavior and elicit knowledge embedded into them during pre-training, not to introduce any new knowledge. Combining activation editing with knowledge updates can be a promising area for future research.

    \item Though HPR outperforms our baselines by a significant margin (i.e., over $15\%$ better than the second best baseline ITI with \texttt{LLama3}), there is still room for improvement. For example, the best MC1 accuracy of HPR on TruthfulQA is currently only about $55\%$ with the base model \texttt{Mistral}. As such, future work can expand our method to develop stronger alignment methods and address safety concerns for LLMs.
    
    %Nevertheless, LLM Safety is a critical issue that must not be overlooked before they can be applied to practical application. As such, improving safety in LLMs and methods to efficiently achieve that should be further studied.

    \item HPR has been shown to perform well on a variety of behavior-related tasks. However, our experiments involves only English data, thus not fully reflecting the capability of the proposed method for multilingual LLMs and data. Future work can explore the effectiveness of our method for multilingual settings, aiming for more robust methods for diverse languages and multilingual LLMs.
    
    %Since Activation Editing methods can change LLMs' behaviour, it would be interesting to see if they can be used to transfer behaviour between languages.

\end{itemize}

\section*{Ethics Statement}
\label{sec:ethics_statement}
Our work utilize open-source LLMs, i.e., \texttt{Llama2-7B-Chat} \citep{touvron2023llama2}, \texttt{Mistral-7B-Instruct} \citep{jiang2023mistral}, and \texttt{Llama3-8B-Instruct} \citep{llama3modelcard}, as the base models, thus potentially inheriting their inherent societal issues like bias, hallucination, privacy, etc. Simultaneously, we propose a novel activation editing method aiming at altering LLMs' behaviour for the better, contributing to the on-going efforts to advance LLM safety. As activation and model editing for LLMs has been studied in recent published work \cite{NEURIPS2023_81b83900,liu-etal-2021-dexperts,ilharco2023editing}, we do not believe our work poses greater societal risks than such studies and open-source LLMs. Finally, we confirm that we follow all the ethical guideline from ACL \iffalse ARR\fi to the best of our knowledge when performing this research.

%Although the proposed method is intended as the means to invoke desirable behaviors, we must acknowledge the potential of it being used by malicious agents for ill intent purposes such as promoting harmful generation and bypassing safety mechanism. Preventing this possibility is an important research direction that we will look into in future study.

\section*{Acknowledgements}

This research has been supported by the Army Research Office (ARO) grant W911NF-21-1-0112, the NSF grant CNS-1747798 to the IUCRC Center for Big Learning, and the NSF grant \# 2239570. This research is also supported in part by the Office of the Director of National Intelligence (ODNI), Intelligence Advanced Research Projects Activity (IARPA), via the HIATUS Program contract 2022-22072200003. The views and conclusions contained herein are those of the authors and should not be interpreted as necessarily representing the official policies, either expressed or implied, of ODNI, IARPA, or the U.S. Government.

\bibliography{anthology,custom}

\begin{thebibliography}{44}
\providecommand{\natexlab}[1]{#1}

\bibitem[{AI@Meta(2024)}]{llama3modelcard}
AI@Meta. 2024.
\newblock \href {https://github.com/meta-llama/llama3/blob/main/MODEL_CARD.md} {Llama 3 model card}.

\bibitem[{Bai et~al.(2022)Bai, Jones, and et~al.}]{bai2022training}
Yuntao Bai, Andy Jones, and et~al. 2022.
\newblock \href {https://arxiv.org/abs/2204.05862} {Training a helpful and harmless assistant with reinforcement learning from human feedback}.
\newblock \emph{Preprint}, arXiv:2204.05862.

\bibitem[{Brown et~al.(2023)Brown, Godfrey, Nizinski, Tu, and Kvinge}]{brown2023robustness}
Davis Brown, Charles Godfrey, Cody Nizinski, Jonathan Tu, and Henry Kvinge. 2023.
\newblock Robustness of edited neural networks.
\newblock In \emph{ICLR 2023 Workshop on Mathematical and Empirical Understanding of Foundation Models}.

\bibitem[{Brown et~al.(2020)Brown, Mann, Ryder, Subbiah, Kaplan, Dhariwal, Neelakantan, Shyam, Sastry, Askell, Agarwal, Herbert-Voss, Krueger, Henighan, Child, Ramesh, Ziegler, Wu, Winter, Hesse, Chen, Sigler, Litwin, Gray, Chess, Clark, Berner, McCandlish, Radford, Sutskever, and Amodei}]{NEURIPS2020_1457c0d6}
Tom Brown, Benjamin Mann, Nick Ryder, Melanie Subbiah, Jared~D Kaplan, Prafulla Dhariwal, Arvind Neelakantan, Pranav Shyam, Girish Sastry, Amanda Askell, Sandhini Agarwal, Ariel Herbert-Voss, Gretchen Krueger, Tom Henighan, Rewon Child, Aditya Ramesh, Daniel Ziegler, Jeffrey Wu, Clemens Winter, Chris Hesse, Mark Chen, Eric Sigler, Mateusz Litwin, Scott Gray, Benjamin Chess, Jack Clark, Christopher Berner, Sam McCandlish, Alec Radford, Ilya Sutskever, and Dario Amodei. 2020.
\newblock \href {https://proceedings.neurips.cc/paper_files/paper/2020/file/1457c0d6bfcb4967418bfb8ac142f64a-Paper.pdf} {Language models are few-shot learners}.
\newblock In \emph{Advances in Neural Information Processing Systems}, volume~33, pages 1877--1901. Curran Associates, Inc.

\bibitem[{Burns et~al.(2023)Burns, Ye, Klein, and Steinhardt}]{burns2023discovering}
Collin Burns, Haotian Ye, Dan Klein, and Jacob Steinhardt. 2023.
\newblock \href {https://openreview.net/forum?id=ETKGuby0hcs} {Discovering latent knowledge in language models without supervision}.
\newblock In \emph{The Eleventh International Conference on Learning Representations}.

\bibitem[{Chowdhery et~al.(2022)Chowdhery, Narang, Devlin, Bosma, Mishra, Roberts, Barham, Chung, Sutton, Gehrmann, Schuh, Shi, Tsvyashchenko, Maynez, Rao, Barnes, Tay, Shazeer, Prabhakaran, Reif, Du, Hutchinson, Pope, Bradbury, Austin, Isard, Gur-Ari, Yin, Duke, Levskaya, Ghemawat, Dev, Michalewski, Garcia, Misra, Robinson, Fedus, Zhou, Ippolito, Luan, Lim, Zoph, Spiridonov, Sepassi, Dohan, Agrawal, Omernick, Dai, Pillai, Pellat, Lewkowycz, Moreira, Child, Polozov, Lee, Zhou, Wang, Saeta, Diaz, Firat, Catasta, Wei, Meier-Hellstern, Eck, Dean, Petrov, and Fiedel}]{chowdhery2022palm}
Aakanksha Chowdhery, Sharan Narang, Jacob Devlin, Maarten Bosma, Gaurav Mishra, Adam Roberts, Paul Barham, Hyung~Won Chung, Charles Sutton, Sebastian Gehrmann, Parker Schuh, Kensen Shi, Sasha Tsvyashchenko, Joshua Maynez, Abhishek Rao, Parker Barnes, Yi~Tay, Noam Shazeer, Vinodkumar Prabhakaran, Emily Reif, Nan Du, Ben Hutchinson, Reiner Pope, James Bradbury, Jacob Austin, Michael Isard, Guy Gur-Ari, Pengcheng Yin, Toju Duke, Anselm Levskaya, Sanjay Ghemawat, Sunipa Dev, Henryk Michalewski, Xavier Garcia, Vedant Misra, Kevin Robinson, Liam Fedus, Denny Zhou, Daphne Ippolito, David Luan, Hyeontaek Lim, Barret Zoph, Alexander Spiridonov, Ryan Sepassi, David Dohan, Shivani Agrawal, Mark Omernick, Andrew~M. Dai, Thanumalayan~Sankaranarayana Pillai, Marie Pellat, Aitor Lewkowycz, Erica Moreira, Rewon Child, Oleksandr Polozov, Katherine Lee, Zongwei Zhou, Xuezhi Wang, Brennan Saeta, Mark Diaz, Orhan Firat, Michele Catasta, Jason Wei, Kathy Meier-Hellstern, Douglas Eck, Jeff Dean, Slav Petrov, and Noah Fiedel. 2022.
\newblock \href {https://arxiv.org/abs/2204.02311} {Palm: Scaling language modeling with pathways}.
\newblock \emph{Preprint}, arXiv:2204.02311.

\bibitem[{Dai et~al.(2022)Dai, Dong, Hao, Sui, Chang, and Wei}]{dai-etal-2022-knowledge}
Damai Dai, Li~Dong, Yaru Hao, Zhifang Sui, Baobao Chang, and Furu Wei. 2022.
\newblock \href {https://doi.org/10.18653/v1/2022.acl-long.581} {Knowledge neurons in pretrained transformers}.
\newblock In \emph{Proceedings of the 60th Annual Meeting of the Association for Computational Linguistics (Volume 1: Long Papers)}, pages 8493--8502, Dublin, Ireland. Association for Computational Linguistics.

\bibitem[{Dathathri et~al.(2020)Dathathri, Madotto, Lan, Hung, Frank, Molino, Yosinski, and Liu}]{Dathathri2020Plug}
Sumanth Dathathri, Andrea Madotto, Janice Lan, Jane Hung, Eric Frank, Piero Molino, Jason Yosinski, and Rosanne Liu. 2020.
\newblock Plug and play language models: A simple approach to controlled text generation.
\newblock In \emph{International Conference on Learning Representations}.

\bibitem[{Dong et~al.(2023)Dong, Li, Dai, Zheng, Wu, Chang, Sun, Xu, Li, and Sui}]{dong2023survey}
Qingxiu Dong, Lei Li, Damai Dai, Ce~Zheng, Zhiyong Wu, Baobao Chang, Xu~Sun, Jingjing Xu, Lei Li, and Zhifang Sui. 2023.
\newblock \href {https://arxiv.org/abs/2301.00234} {A survey on in-context learning}.
\newblock \emph{Preprint}, arXiv:2301.00234.

\bibitem[{Gao et~al.(2023)Gao, Tow, Abbasi, Biderman, Black, DiPofi, Foster, Golding, Hsu, Le~Noac'h, Li, McDonell, Muennighoff, Ociepa, Phang, Reynolds, Schoelkopf, Skowron, Sutawika, Tang, Thite, Wang, Wang, and Zou}]{eval-harness}
Leo Gao, Jonathan Tow, Baber Abbasi, Stella Biderman, Sid Black, Anthony DiPofi, Charles Foster, Laurence Golding, Jeffrey Hsu, Alain Le~Noac'h, Haonan Li, Kyle McDonell, Niklas Muennighoff, Chris Ociepa, Jason Phang, Laria Reynolds, Hailey Schoelkopf, Aviya Skowron, Lintang Sutawika, Eric Tang, Anish Thite, Ben Wang, Kevin Wang, and Andy Zou. 2023.
\newblock \href {https://doi.org/10.5281/zenodo.10256836} {A framework for few-shot language model evaluation}.

\bibitem[{Hartvigsen et~al.(2022)Hartvigsen, Gabriel, Palangi, Sap, Ray, and Kamar}]{hartvigsen-etal-2022-toxigen}
Thomas Hartvigsen, Saadia Gabriel, Hamid Palangi, Maarten Sap, Dipankar Ray, and Ece Kamar. 2022.
\newblock \href {https://doi.org/10.18653/v1/2022.acl-long.234} {{T}oxi{G}en: A large-scale machine-generated dataset for adversarial and implicit hate speech detection}.
\newblock In \emph{Proceedings of the 60th Annual Meeting of the Association for Computational Linguistics (Volume 1: Long Papers)}, pages 3309--3326, Dublin, Ireland. Association for Computational Linguistics.

\bibitem[{Hernandez et~al.(2023)Hernandez, Li, and Andreas}]{hernandez2023remedi}
Evan Hernandez, Belinda~Z. Li, and Jacob Andreas. 2023.
\newblock Inspecting and editing knowledge representations in language models.
\newblock In \emph{Arxiv}.

\bibitem[{Householder(1958)}]{10.1145/320941.320947}
Alston~S. Householder. 1958.
\newblock \href {https://doi.org/10.1145/320941.320947} {Unitary triangularization of a nonsymmetric matrix}.
\newblock \emph{J. ACM}, 5(4):339–342.

\bibitem[{Hu et~al.(2022)Hu, Shen, Wallis, Allen-Zhu, Li, Wang, Wang, and Chen}]{hu2022lora}
Edward~J Hu, Yelong Shen, Phillip Wallis, Zeyuan Allen-Zhu, Yuanzhi Li, Shean Wang, Lu~Wang, and Weizhu Chen. 2022.
\newblock \href {https://openreview.net/forum?id=nZeVKeeFYf9} {Lo{RA}: Low-rank adaptation of large language models}.
\newblock In \emph{International Conference on Learning Representations}.

\bibitem[{Ilharco et~al.(2023)Ilharco, Ribeiro, Wortsman, Schmidt, Hajishirzi, and Farhadi}]{ilharco2023editing}
Gabriel Ilharco, Marco~Tulio Ribeiro, Mitchell Wortsman, Ludwig Schmidt, Hannaneh Hajishirzi, and Ali Farhadi. 2023.
\newblock Editing models with task arithmetic.
\newblock In \emph{The Eleventh International Conference on Learning Representations}.

\bibitem[{Jiang et~al.(2023)Jiang, Sablayrolles, Mensch, Bamford, Chaplot, de~las Casas, Bressand, Lengyel, Lample, Saulnier, Lavaud, Lachaux, Stock, Scao, Lavril, Wang, Lacroix, and Sayed}]{jiang2023mistral}
Albert~Q. Jiang, Alexandre Sablayrolles, Arthur Mensch, Chris Bamford, Devendra~Singh Chaplot, Diego de~las Casas, Florian Bressand, Gianna Lengyel, Guillaume Lample, Lucile Saulnier, Lélio~Renard Lavaud, Marie-Anne Lachaux, Pierre Stock, Teven~Le Scao, Thibaut Lavril, Thomas Wang, Timothée Lacroix, and William~El Sayed. 2023.
\newblock \href {https://arxiv.org/abs/2310.06825} {Mistral 7b}.
\newblock \emph{Preprint}, arXiv:2310.06825.

\bibitem[{Joshi et~al.(2024)Joshi, Rando, Saparov, Kim, and He}]{joshi2024personas}
Nitish Joshi, Javier Rando, Abulhair Saparov, Najoung Kim, and He~He. 2024.
\newblock \href {https://arxiv.org/abs/2310.18168} {Personas as a way to model truthfulness in language models}.
\newblock \emph{Preprint}, arXiv:2310.18168.

\bibitem[{Krause et~al.(2021)Krause, Gotmare, McCann, Keskar, Joty, Socher, and Rajani}]{krause-etal-2021-gedi-generative}
Ben Krause, Akhilesh~Deepak Gotmare, Bryan McCann, Nitish~Shirish Keskar, Shafiq Joty, Richard Socher, and Nazneen~Fatema Rajani. 2021.
\newblock \href {https://doi.org/10.18653/v1/2021.findings-emnlp.424} {{G}e{D}i: Generative discriminator guided sequence generation}.
\newblock In \emph{Findings of the Association for Computational Linguistics: EMNLP 2021}, pages 4929--4952, Punta Cana, Dominican Republic. Association for Computational Linguistics.

\bibitem[{Lester et~al.(2021)Lester, Al-Rfou, and Constant}]{lester-etal-2021-power}
Brian Lester, Rami Al-Rfou, and Noah Constant. 2021.
\newblock \href {https://doi.org/10.18653/v1/2021.emnlp-main.243} {The power of scale for parameter-efficient prompt tuning}.
\newblock In \emph{Proceedings of the 2021 Conference on Empirical Methods in Natural Language Processing}, pages 3045--3059, Online and Punta Cana, Dominican Republic. Association for Computational Linguistics.

\bibitem[{Li et~al.(2023{\natexlab{a}})Li, Hopkins, Bau, Vi{\'e}gas, Pfister, and Wattenberg}]{li2023emergent}
Kenneth Li, Aspen~K Hopkins, David Bau, Fernanda Vi{\'e}gas, Hanspeter Pfister, and Martin Wattenberg. 2023{\natexlab{a}}.
\newblock Emergent world representations: Exploring a sequence model trained on a synthetic task.
\newblock In \emph{The Eleventh International Conference on Learning Representations}.

\bibitem[{Li et~al.(2023{\natexlab{b}})Li, Patel, Vi\'{e}gas, Pfister, and Wattenberg}]{NEURIPS2023_81b83900}
Kenneth Li, Oam Patel, Fernanda Vi\'{e}gas, Hanspeter Pfister, and Martin Wattenberg. 2023{\natexlab{b}}.
\newblock \href {https://proceedings.neurips.cc/paper_files/paper/2023/file/81b8390039b7302c909cb769f8b6cd93-Paper-Conference.pdf} {Inference-time intervention: Eliciting truthful answers from a language model}.
\newblock In \emph{Advances in Neural Information Processing Systems}, volume~36, pages 41451--41530. Curran Associates, Inc.

\bibitem[{Li et~al.(2023{\natexlab{c}})Li, Holtzman, Fried, Liang, Eisner, Hashimoto, Zettlemoyer, and Lewis}]{li-etal-2023-contrastive}
Xiang~Lisa Li, Ari Holtzman, Daniel Fried, Percy Liang, Jason Eisner, Tatsunori Hashimoto, Luke Zettlemoyer, and Mike Lewis. 2023{\natexlab{c}}.
\newblock \href {https://doi.org/10.18653/v1/2023.acl-long.687} {Contrastive decoding: Open-ended text generation as optimization}.
\newblock In \emph{Proceedings of the 61st Annual Meeting of the Association for Computational Linguistics (Volume 1: Long Papers)}, pages 12286--12312, Toronto, Canada. Association for Computational Linguistics.

\bibitem[{Li and Liang(2021)}]{li-liang-2021-prefix}
Xiang~Lisa Li and Percy Liang. 2021.
\newblock \href {https://doi.org/10.18653/v1/2021.acl-long.353} {Prefix-tuning: Optimizing continuous prompts for generation}.
\newblock In \emph{Proceedings of the 59th Annual Meeting of the Association for Computational Linguistics and the 11th International Joint Conference on Natural Language Processing (Volume 1: Long Papers)}, pages 4582--4597, Online. Association for Computational Linguistics.

\bibitem[{Lin et~al.(2022)Lin, Hilton, and Evans}]{lin-etal-2022-truthfulqa}
Stephanie Lin, Jacob Hilton, and Owain Evans. 2022.
\newblock \href {https://doi.org/10.18653/v1/2022.acl-long.229} {{T}ruthful{QA}: Measuring how models mimic human falsehoods}.
\newblock In \emph{Proceedings of the 60th Annual Meeting of the Association for Computational Linguistics (Volume 1: Long Papers)}, pages 3214--3252, Dublin, Ireland. Association for Computational Linguistics.

\bibitem[{Liu et~al.(2024)Liu, Han, Wang, Tsvetkov, Choi, and Smith}]{liu2024tuning}
Alisa Liu, Xiaochuang Han, Yizhong Wang, Yulia Tsvetkov, Yejin Choi, and Noah~A. Smith. 2024.
\newblock \href {https://arxiv.org/abs/2401.08565} {Tuning language models by proxy}.
\newblock \emph{Preprint}, arXiv:2401.08565.

\bibitem[{Liu et~al.(2021)Liu, Sap, Lu, Swayamdipta, Bhagavatula, Smith, and Choi}]{liu-etal-2021-dexperts}
Alisa Liu, Maarten Sap, Ximing Lu, Swabha Swayamdipta, Chandra Bhagavatula, Noah~A. Smith, and Yejin Choi. 2021.
\newblock \href {https://doi.org/10.18653/v1/2021.acl-long.522} {{DE}xperts: Decoding-time controlled text generation with experts and anti-experts}.
\newblock In \emph{Proceedings of the 59th Annual Meeting of the Association for Computational Linguistics and the 11th International Joint Conference on Natural Language Processing (Volume 1: Long Papers)}, pages 6691--6706, Online. Association for Computational Linguistics.

\bibitem[{Loshchilov and Hutter(2019)}]{loshchilov2018decoupled}
Ilya Loshchilov and Frank Hutter. 2019.
\newblock \href {https://openreview.net/forum?id=Bkg6RiCqY7} {Decoupled weight decay regularization}.
\newblock In \emph{International Conference on Learning Representations}.

\bibitem[{Meng et~al.(2022)Meng, Bau, Andonian, and Belinkov}]{meng2022locating}
Kevin Meng, David Bau, Alex Andonian, and Yonatan Belinkov. 2022.
\newblock Locating and editing factual associations in gpt.
\newblock In \emph{Advances in Neural Information Processing Systems}.

\bibitem[{Merity et~al.(2017)Merity, Xiong, Bradbury, and Socher}]{merity2017pointer}
Stephen Merity, Caiming Xiong, James Bradbury, and Richard Socher. 2017.
\newblock \href {https://openreview.net/forum?id=Byj72udxe} {Pointer sentinel mixture models}.
\newblock In \emph{International Conference on Learning Representations}.

\bibitem[{Mikolov et~al.(2013)Mikolov, Yih, and Zweig}]{mikolov-etal-2013-linguistic}
Tomas Mikolov, Wen-tau Yih, and Geoffrey Zweig. 2013.
\newblock \href {https://aclanthology.org/N13-1090} {Linguistic regularities in continuous space word representations}.
\newblock In \emph{Proceedings of the 2013 Conference of the North {A}merican Chapter of the Association for Computational Linguistics: Human Language Technologies}, pages 746--751, Atlanta, Georgia. Association for Computational Linguistics.

\bibitem[{OpenAI(2024)}]{openai2024gpt4}
OpenAI. 2024.
\newblock \href {https://arxiv.org/abs/2303.08774} {Gpt-4 technical report}.
\newblock \emph{Preprint}, arXiv:2303.08774.

\bibitem[{Ouyang et~al.(2022)Ouyang, Wu, Jiang, Almeida, Wainwright, Mishkin, Zhang, Agarwal, Slama, Ray, Schulman, Hilton, Kelton, Miller, Simens, Askell, Welinder, Christiano, Leike, and Lowe}]{NEURIPS2022_b1efde53}
Long Ouyang, Jeffrey Wu, Xu~Jiang, Diogo Almeida, Carroll Wainwright, Pamela Mishkin, Chong Zhang, Sandhini Agarwal, Katarina Slama, Alex Ray, John Schulman, Jacob Hilton, Fraser Kelton, Luke Miller, Maddie Simens, Amanda Askell, Peter Welinder, Paul~F Christiano, Jan Leike, and Ryan Lowe. 2022.
\newblock \href {https://proceedings.neurips.cc/paper_files/paper/2022/file/b1efde53be364a73914f58805a001731-Paper-Conference.pdf} {Training language models to follow instructions with human feedback}.
\newblock In \emph{Advances in Neural Information Processing Systems}, volume~35, pages 27730--27744. Curran Associates, Inc.

\bibitem[{Parrish et~al.(2022)Parrish, Chen, Nangia, Padmakumar, Phang, Thompson, Htut, and Bowman}]{parrish-etal-2022-bbq}
Alicia Parrish, Angelica Chen, Nikita Nangia, Vishakh Padmakumar, Jason Phang, Jana Thompson, Phu~Mon Htut, and Samuel Bowman. 2022.
\newblock \href {https://doi.org/10.18653/v1/2022.findings-acl.165} {{BBQ}: A hand-built bias benchmark for question answering}.
\newblock In \emph{Findings of the Association for Computational Linguistics: ACL 2022}, pages 2086--2105, Dublin, Ireland. Association for Computational Linguistics.

\bibitem[{Qi et~al.(2023)Qi, Zeng, Xie, Chen, Jia, Mittal, and Henderson}]{qi2023finetuning}
Xiangyu Qi, Yi~Zeng, Tinghao Xie, Pin-Yu Chen, Ruoxi Jia, Prateek Mittal, and Peter Henderson. 2023.
\newblock \href {https://arxiv.org/abs/2310.03693} {Fine-tuning aligned language models compromises safety, even when users do not intend to!}
\newblock \emph{Preprint}, arXiv:2310.03693.

\bibitem[{Radford and Narasimhan(2018)}]{Radford2018ImprovingLU}
Alec Radford and Karthik Narasimhan. 2018.
\newblock Improving language understanding by generative pre-training.

\bibitem[{Radford et~al.(2019)Radford, Wu, Child, Luan, Amodei, and Sutskever}]{radford2019language}
Alec Radford, Jeff Wu, Rewon Child, David Luan, Dario Amodei, and Ilya Sutskever. 2019.
\newblock Language models are unsupervised multitask learners.

\bibitem[{Rimsky et~al.(2024)Rimsky, Gabrieli, Schulz, Tong, Hubinger, and Turner}]{rimsky2024steering}
Nina Rimsky, Nick Gabrieli, Julian Schulz, Meg Tong, Evan Hubinger, and Alexander~Matt Turner. 2024.
\newblock \href {https://arxiv.org/abs/2312.06681} {Steering llama 2 via contrastive activation addition}.
\newblock \emph{Preprint}, arXiv:2312.06681.

\bibitem[{Srivastava et~al.(2023)Srivastava, Rastogi, and et~al.}]{srivastava2023beyond}
Aarohi Srivastava, Abhinav Rastogi, and et~al. 2023.
\newblock \href {https://openreview.net/forum?id=uyTL5Bvosj} {Beyond the imitation game: Quantifying and extrapolating the capabilities of language models}.
\newblock \emph{Transactions on Machine Learning Research}.

\bibitem[{Touvron et~al.(2023{\natexlab{a}})Touvron, Lavril, Izacard, Martinet, Lachaux, Lacroix, Rozière, Goyal, Hambro, Azhar, Rodriguez, Joulin, Grave, and Lample}]{touvron2023llama}
Hugo Touvron, Thibaut Lavril, Gautier Izacard, Xavier Martinet, Marie-Anne Lachaux, Timothée Lacroix, Baptiste Rozière, Naman Goyal, Eric Hambro, Faisal Azhar, Aurelien Rodriguez, Armand Joulin, Edouard Grave, and Guillaume Lample. 2023{\natexlab{a}}.
\newblock \href {https://arxiv.org/abs/2302.13971} {Llama: Open and efficient foundation language models}.
\newblock \emph{Preprint}, arXiv:2302.13971.

\bibitem[{Touvron et~al.(2023{\natexlab{b}})Touvron, Martin, Stone, Albert, Almahairi, Babaei, Bashlykov, Batra, Bhargava, Bhosale, Bikel, Blecher, Ferrer, Chen, Cucurull, Esiobu, Fernandes, Fu, Fu, Fuller, Gao, Goswami, Goyal, Hartshorn, Hosseini, Hou, Inan, Kardas, Kerkez, Khabsa, Kloumann, Korenev, Koura, Lachaux, Lavril, Lee, Liskovich, Lu, Mao, Martinet, Mihaylov, Mishra, Molybog, Nie, Poulton, Reizenstein, Rungta, Saladi, Schelten, Silva, Smith, Subramanian, Tan, Tang, Taylor, Williams, Kuan, Xu, Yan, Zarov, Zhang, Fan, Kambadur, Narang, Rodriguez, Stojnic, Edunov, and Scialom}]{touvron2023llama2}
Hugo Touvron, Louis Martin, Kevin Stone, Peter Albert, Amjad Almahairi, Yasmine Babaei, Nikolay Bashlykov, Soumya Batra, Prajjwal Bhargava, Shruti Bhosale, Dan Bikel, Lukas Blecher, Cristian~Canton Ferrer, Moya Chen, Guillem Cucurull, David Esiobu, Jude Fernandes, Jeremy Fu, Wenyin Fu, Brian Fuller, Cynthia Gao, Vedanuj Goswami, Naman Goyal, Anthony Hartshorn, Saghar Hosseini, Rui Hou, Hakan Inan, Marcin Kardas, Viktor Kerkez, Madian Khabsa, Isabel Kloumann, Artem Korenev, Punit~Singh Koura, Marie-Anne Lachaux, Thibaut Lavril, Jenya Lee, Diana Liskovich, Yinghai Lu, Yuning Mao, Xavier Martinet, Todor Mihaylov, Pushkar Mishra, Igor Molybog, Yixin Nie, Andrew Poulton, Jeremy Reizenstein, Rashi Rungta, Kalyan Saladi, Alan Schelten, Ruan Silva, Eric~Michael Smith, Ranjan Subramanian, Xiaoqing~Ellen Tan, Binh Tang, Ross Taylor, Adina Williams, Jian~Xiang Kuan, Puxin Xu, Zheng Yan, Iliyan Zarov, Yuchen Zhang, Angela Fan, Melanie Kambadur, Sharan Narang, Aurelien Rodriguez, Robert Stojnic, Sergey Edunov, and Thomas
  Scialom. 2023{\natexlab{b}}.
\newblock \href {https://arxiv.org/abs/2307.09288} {Llama 2: Open foundation and fine-tuned chat models}.
\newblock \emph{Preprint}, arXiv:2307.09288.

\bibitem[{Turner et~al.(2023)Turner, Thiergart, Udell, Leech, Mini, and MacDiarmid}]{turner2023activation}
Alexander~Matt Turner, Lisa Thiergart, David Udell, Gavin Leech, Ulisse Mini, and Monte MacDiarmid. 2023.
\newblock \href {https://arxiv.org/abs/2308.10248} {Activation addition: Steering language models without optimization}.
\newblock \emph{Preprint}, arXiv:2308.10248.

\bibitem[{von Rütte et~al.(2024)von Rütte, Anagnostidis, Bachmann, and Hofmann}]{vonrütte2024language}
Dimitri von Rütte, Sotiris Anagnostidis, Gregor Bachmann, and Thomas Hofmann. 2024.
\newblock \href {https://arxiv.org/abs/2402.14433} {A language model's guide through latent space}.
\newblock \emph{Preprint}, arXiv:2402.14433.

\bibitem[{Wan et~al.(2024)Wan, Wang, Liu, Alam, Zheng, Liu, Qu, Yan, Zhu, Zhang, Chowdhury, and Zhang}]{wan2024efficient}
Zhongwei Wan, Xin Wang, Che Liu, Samiul Alam, Yu~Zheng, Jiachen Liu, Zhongnan Qu, Shen Yan, Yi~Zhu, Quanlu Zhang, Mosharaf Chowdhury, and Mi~Zhang. 2024.
\newblock \href {https://openreview.net/forum?id=bsCCJHbO8A} {Efficient large language models: A survey}.
\newblock \emph{Transactions on Machine Learning Research}.
\newblock Survey Certification.

\bibitem[{Wei et~al.(2022)Wei, Tay, Bommasani, Raffel, Zoph, Borgeaud, Yogatama, Bosma, Zhou, Metzler, Chi, Hashimoto, Vinyals, Liang, Dean, and Fedus}]{wei2022emergent}
Jason Wei, Yi~Tay, Rishi Bommasani, Colin Raffel, Barret Zoph, Sebastian Borgeaud, Dani Yogatama, Maarten Bosma, Denny Zhou, Donald Metzler, Ed~H. Chi, Tatsunori Hashimoto, Oriol Vinyals, Percy Liang, Jeff Dean, and William Fedus. 2022.
\newblock \href {https://openreview.net/forum?id=yzkSU5zdwD} {Emergent abilities of large language models}.
\newblock \emph{Transactions on Machine Learning Research}.
\newblock Survey Certification.

\end{thebibliography}
% Custom bibliography entries only
% \bibliography{custom}

\clearpage

\appendix

\section{Derivation of Equation \ref{eq:rotated_activation}}
\label{sec:appendix_proof}

\begin{figure*}[hbt!]
    \centering
    \begin{subfigure}{0.48\textwidth}
        \includegraphics[width=\linewidth]{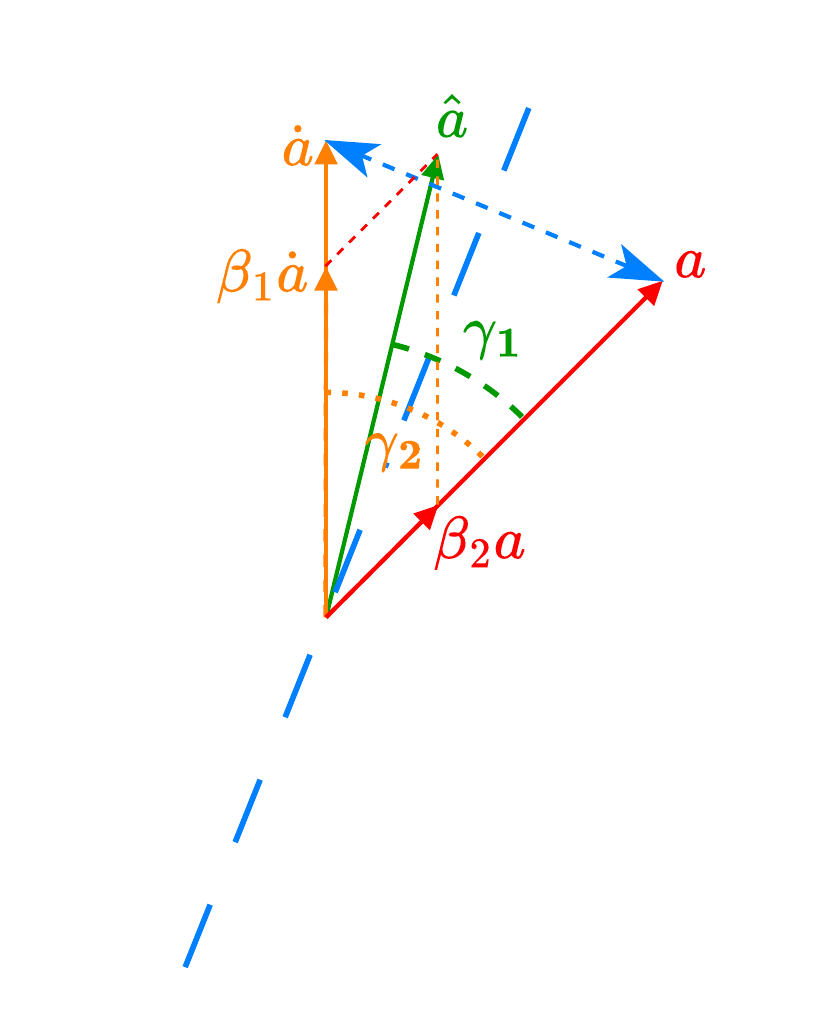}
        \caption{First case: $\gamma_1 < \gamma_2$}
        \label{fig:proof_case_1}
    \end{subfigure}
    \begin{subfigure}{0.48\textwidth}
        \includegraphics[width=\linewidth]{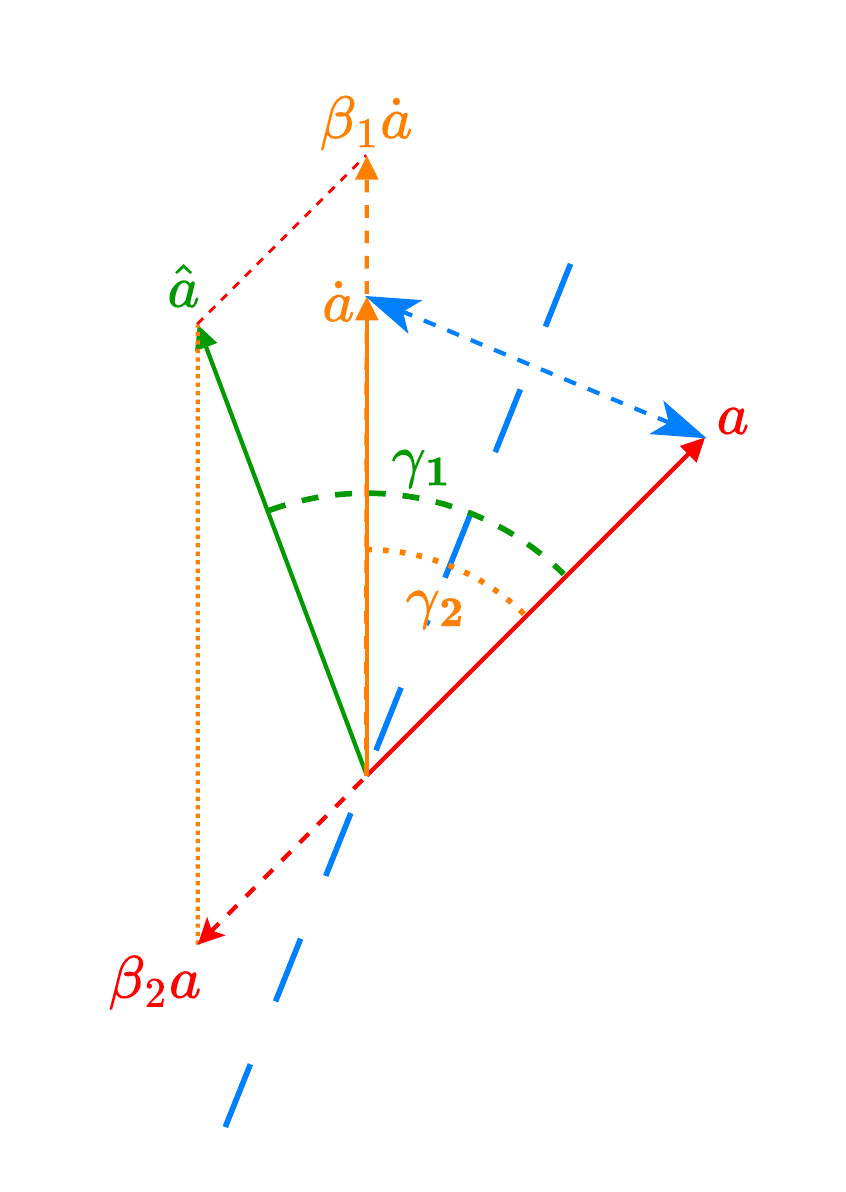}
        \caption{Second case: $\gamma_1 > \gamma_2$}
        \label{fig:proof_case_2}
    \end{subfigure}
    \caption{Illustration of the two cases when rotating vector in $2$-D plane.}
    \label{fig:rotated_act_proof}
\end{figure*}

In this section we describe the process of deriving Equation \ref{eq:rotated_activation}. Since the rotation of interest occurs on a $2$-D plane, and $\lVert \hat{a} \rVert = \lVert \dot{a} \rVert = \lVert a \rVert$, we can calculate $\hat{a}$ by combining $a$ and $\dot{a}$. If $\gamma_1 = \gamma_2$, Equation \ref{eq:rotated_activation} trivially holds: $\hat{a} = \dot{a}$.  If not, there are two cases that can occur: $\gamma_1 < \gamma_2$, and $\gamma_1 > \gamma_2$. We illustrate both of them in Figure \ref{fig:rotated_act_proof} to make the derivation easier to follow. In this figure, we color the original negative activation $a$ in \textcolor{red}{red}, the target positive activation $\hat{a}$ in \textcolor{green}{green}, and the intermediate vector $\dot{a}$ in \textcolor{orange}{orange}. 

Say, we have
\begin{equation}
    \hat{a} = \beta_1 \dot{a} + \beta_2 a
    \label{rotated_activation_prototype}
\end{equation}

In the first case (Figure \ref{fig:proof_case_1}), applying the law of sines in trigonometry, we obtain
\begin{equation}
    \frac{\lVert \hat{a} \rVert}{\sin(\pi - \gamma_2)}
    = \frac{\beta_1 \lVert \dot{a} \rVert}{\sin(\gamma_1)}
    = \frac{\beta_2 \lVert a \rVert}{\sin(\gamma_2 - \gamma_1)}
\end{equation}
This is equivalent to
\begin{equation}
    \frac{1}{\sin(\gamma_2)}
    = \frac{\beta_1}{\sin(\gamma_1)}
    = \frac{\beta_2}{\sin(\gamma_2 - \gamma_1)}
\end{equation}
Thus,
\begin{equation}
    \beta_1 = \frac{\sin(\gamma_1)}{\sin(\gamma_2)}
\end{equation}
\begin{equation}
    \beta_2 = \frac{\sin(\gamma_2 - \gamma_1)}{\sin(\gamma_2)}
\end{equation}

Similarly for the second case (Figure \ref{fig:proof_case_2}), we have
\begin{equation}
    \frac{1}{\sin(\gamma_2)}
    = \frac{\beta_1}{\sin(\pi - \gamma_1)}
    = \frac{-\beta_2}{\sin(\gamma_1 - \gamma_2)}
\end{equation}
\begin{equation}
    \implies \frac{1}{\sin(\gamma_2)}
    = \frac{\beta_1}{\sin(\gamma_1)}
    = \frac{\beta_2}{\sin(\gamma_2 - \gamma_1)}
\end{equation}
\begin{equation}
    \implies
    \begin{cases}
        \beta_1 = \frac{\sin(\gamma_1)}{\sin(\gamma_2)} \\
        \beta_2 = \frac{\sin(\gamma_2 - \gamma_1)}{\sin(\gamma_2)}
    \end{cases}
\end{equation}

Combining both cases, we arrive at a general formula for calculating the target activation vector:
\begin{equation}
    \centering
    \hat{a} = \frac{\sin(\gamma_1)}{\sin(\gamma_2)} \dot{a} + \frac{\sin(\gamma_2 - \gamma_1)}{\sin(\gamma_2)} a
\end{equation}

% \section{Additional Details about Experiments}
% \label{sec:appendix_exp_details}

\begin{figure*}[hbt!]
    \begin{subfigure}{0.33\textwidth}
        \includegraphics[width=\linewidth]{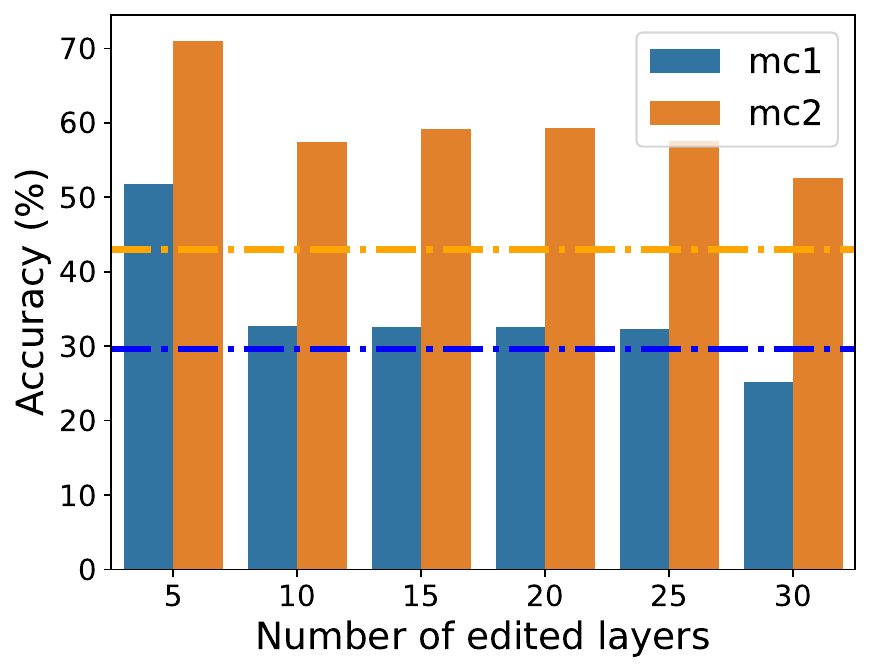}
        \caption {\texttt{Llama2-7B-Chat}}
        \label{fig:num_layers_llama2}
    \end{subfigure}
    \begin{subfigure}{0.33\textwidth}
        \includegraphics[width=\linewidth]{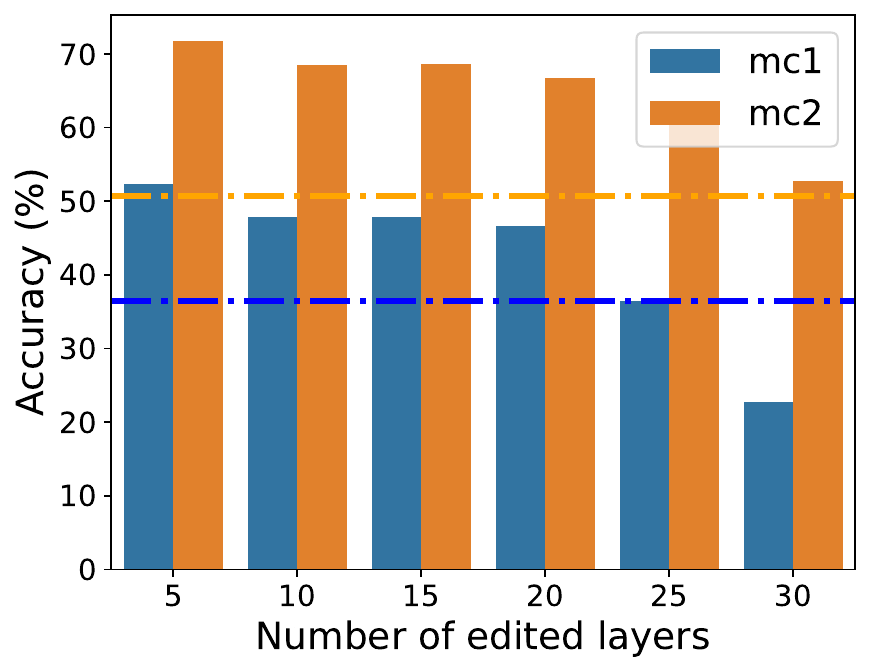}
        \caption {\texttt{Llama3-8B-Instruct}}
        \label{fig:num_layers_llama3}
    \end{subfigure}
    \begin{subfigure}{0.33\textwidth}
        \includegraphics[width=\linewidth]{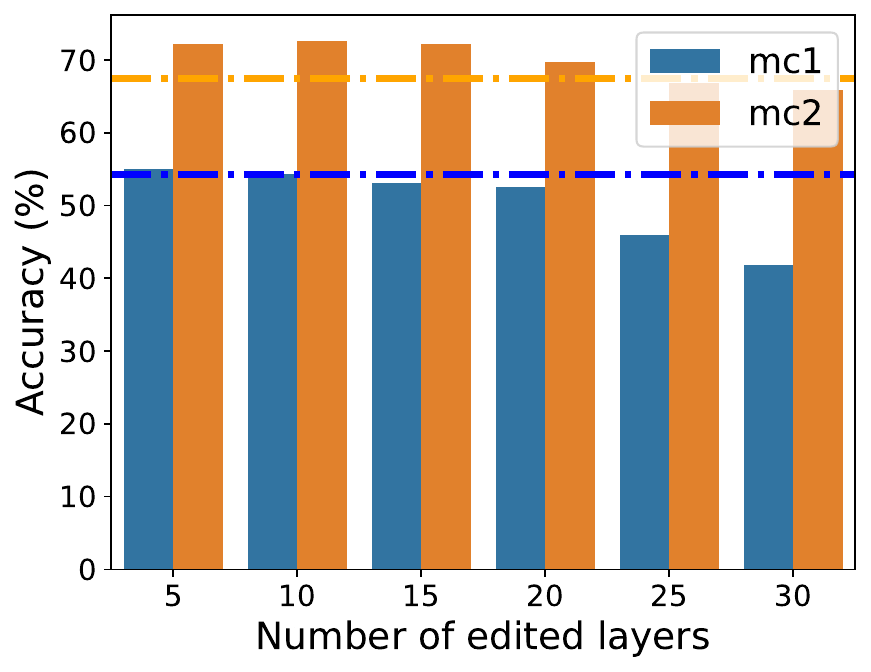}
        \caption {\texttt{Mistral-7B-Instruct}}
        \label{fig:num_layers_mistral}
    \end{subfigure}

    \caption{HPR's performance on TruthfulQA with different numbers of edited layers.}
    \label{fig:num_edited_layers}
\end{figure*}

\section{Training efficiency}
\label{sec:appendix_training_efficiency}
During the training phase, we use $a_{i,j}^{(l), \mathbf{p}}$ / $a_{i,j}^{(l), \mathbf{n}}$ pairs to form the inputs and labels for the linear probe and angle prediction modules in each layer. Generally, these are computed by passing training data samples $x_i \Vert y_i^{\mathbf{p}}$ and $x_i \Vert y_i^{\mathbf{n}}$ through the model $\mathcal{M}$ and record the activations at each layer and token position. However, since our method does not update the parameters of $\mathcal{M}$, its activation vectors can be treated as constants. Thus, before training we pre-compute all activations on the training data to make a dataset of $a_{i,j}^{(l), \mathbf{p}}$ / $a_{i,j}^{(l), \mathbf{n}}$ pairs for each layer. These can then be used to train the linear probe and angle prediction modules independently of the base model. In this way, the base LLM does not need to be loaded into GPU RAM, saving more space for training the HPR modules.

\section{Evaluating Different Numbers of Editted Layers}
\label{sec:appendix_num_layers}

Motivated by the varying linear probing accuracy across different layers in LLMs for positive and negative activations in Figure \ref{fig:probe_acc_llama2}, our method HPR choose the top $k$ layers with highest probe accuracy in LLMs for activation editing. Figure \ref{fig:num_edited_layers} illustrates the performance of HPR using different values of $k$ for all the three base LLMs. The bars depict MC1 (blue) and MC2 (orange) accuracy. We also add the performance of the respective base LLM and illustrate them with horizontal lines for comparison. It is clear from the figure that editing only the top $5$ layers yields the best performance across models. As we increase the number of edited layers, multiple choice accuracy decreases, even falling below baseline in the case of \texttt{Mistral-7B-Instruct}. This can be partly attributed to aggregated error from imperfect linear probes (Figure \ref{fig:probe_acc_llama2}).

\end{document}